\def\paperID{9494}
\def\confName{ICCV}
\def\confYear{2025}

\def\paperTitle{LongSplat: Robust Unposed 3D Gaussian Splatting for Casual Long Videos}

\def\authorBlock{
    Chin-Yang Lin$^{1}$
    \quad
    Cheng Sun$^{2}$
    \quad
    Fu-En Yang$^{2}$
    \\
    Min-Hung Chen$^{2}$
    \quad
    Yen-Yu Lin$^{1}$
    \quad
     Yu-Lun Liu$^{1}$\vspace{0.5em}
    \\
    \centerline{$^1$National Yang Ming Chiao Tung University \quad $^2$NVIDIA Research}
}

\newif\ifreview 
\newif\ifarxiv \newcommand{\arxiv}{\arxivtrue}
\newif\ifcamera 
\newif\ifrebuttal 

\arxiv 

\pdfoutput=1
\documentclass[10pt,twocolumn,letterpaper]{article}
\ifreview \usepackage[review]{cvpr} \fi
\ifarxiv \usepackage[pagenumbers]{cvpr} \fi
\ifrebuttal \usepackage[rebuttal]{cvpr} \fi
\ifcamera \usepackage{cvpr} \fi

\usepackage{graphicx}	
\usepackage{amsmath}	
\usepackage{amssymb}	
\usepackage{booktabs}
\usepackage{times}
\usepackage{microtype}
\usepackage{epsfig}
\usepackage{caption}
\usepackage{float}
\usepackage{placeins}
\usepackage{color, colortbl}
\usepackage{stfloats}
\usepackage{enumitem}
\usepackage{tabularx}
\usepackage{xstring}
\usepackage{multirow}
\usepackage{xspace}
\usepackage{url}
\usepackage{subcaption}
\usepackage{xcolor}
\usepackage[hang,flushmargin]{footmisc}
\usepackage[ruled,vlined]{algorithm2e}
\usepackage[accsupp]{axessibility}

\ifcamera \usepackage[accsupp]{axessibility} \fi

% ...

\ifarxiv  \fi

\newcommand{\R}[1]{{%
    \textbf{%
        \ifstrequal{#1}{1}{\textcolor{red}{R#1}}{%
        \ifstrequal{#1}{2}{\textcolor{blue}{R#1}}{%
        \ifstrequal{#1}{3}{\textcolor{magenta}{R#1}}{%
        \ifstrequal{#1}{4}{\textcolor{teal}{R#1}}{%
                           \textcolor{cyan}{R#1}%
        }}}}%
    }%
}}

\usepackage{xr-hyper}

\makeatletter
\newcommand*{\addFileDependency}[1]{
  \typeout{(#1)}
  \@addtofilelist{#1}
  \IfFileExists{#1}{}{\typeout{No file #1.}}
}

\makeatother
\newcommand*{\myexternaldocument}[1]{
    \externaldocument{#1}
    \addFileDependency{#1.tex}
    \addFileDependency{#1.aux}
}

\definecolor{cvprblue}{rgb}{0.21,0.49,0.74}
\usepackage[pagebackref,breaklinks,colorlinks,allcolors=cvprblue]{hyperref}
\usepackage[capitalize]{cleveref}
\crefname{section}{Sec.}{Secs.}
\crefname{table}{Table}{Tables}
\crefname{figure}{Fig.}{Figs.}

\ifarxiv \crefname{appendix}{App.}{Apps.}
\else \crefname{appendix}{Suppl.}{Suppls.} \fi

\unless\ifarxiv \myexternaldocument{_supplementary} \fi

\begin{document}
\title{\paperTitle}
\author{\authorBlock}

\twocolumn[{%
\renewcommand\twocolumn[1][]{#1}%
\maketitle
\begin{center}
\centering
\vspace{-5mm}
\captionsetup{type=figure}
\resizebox{1.0\textwidth}{!} 
{
\includegraphics[width=\textwidth]{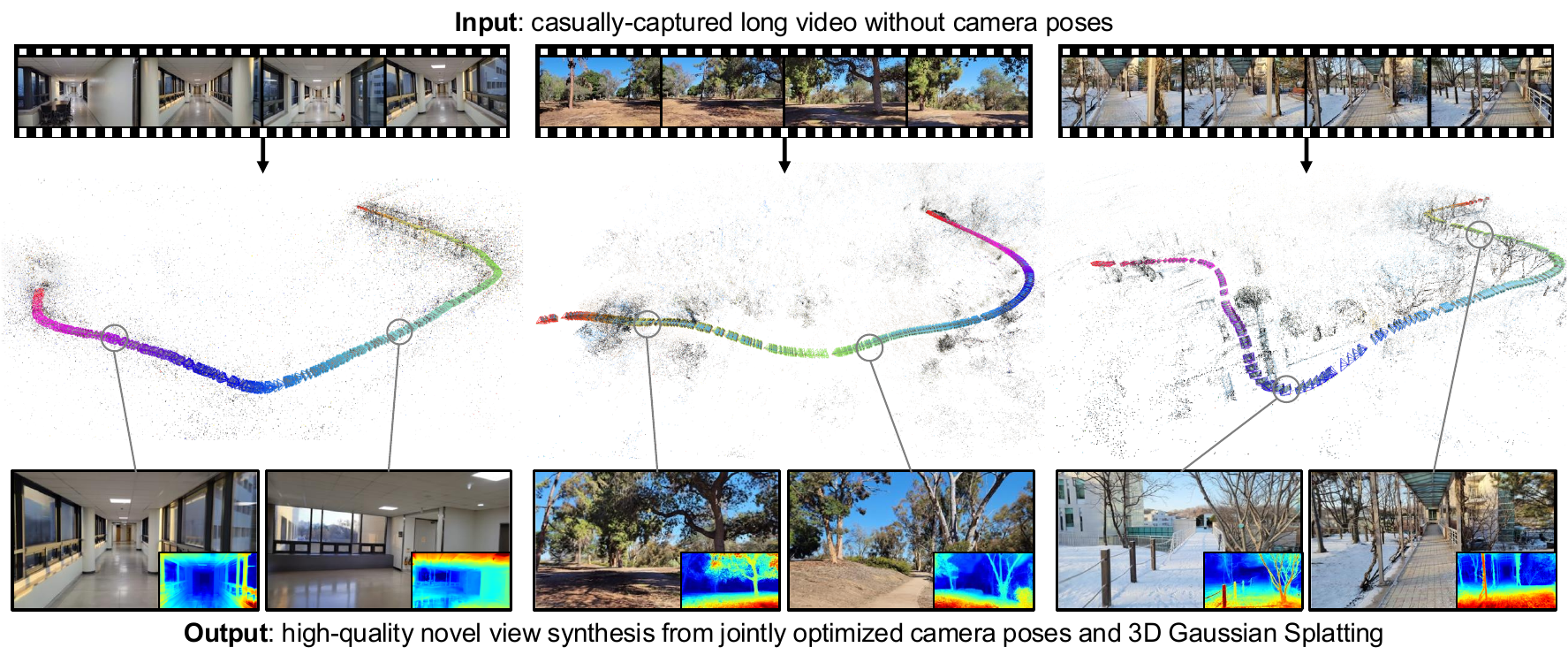}
}
\vspace{-6mm}
\caption{
\textbf{LongSplat achieves robust novel view synthesis from casually captured long videos without provided camera poses.} 
Our approach jointly optimizes camera poses and 3D Gaussian Splatting, producing accurate and visually coherent reconstructions even under challenging conditions.
}
\label{teaser}
\end{center}
}]

\maketitle

\begin{abstract}
LongSplat addresses critical challenges in novel view synthesis (NVS) from casually captured long videos characterized by irregular camera motion, unknown camera poses, and expansive scenes. Current methods often suffer from pose drift, inaccurate geometry initialization, and severe memory limitations. To address these issues, we introduce LongSplat, a robust unposed 3D Gaussian Splatting framework featuring: (1) Incremental Joint Optimization that concurrently optimizes camera poses and 3D Gaussians to avoid local minima and ensure global consistency; (2) a robust Pose Estimation Module leveraging learned 3D priors; and (3) an efficient Octree Anchor Formation mechanism that converts dense point clouds into anchors based on spatial density. Extensive experiments on challenging benchmarks demonstrate that LongSplat achieves state-of-the-art results, substantially improving rendering quality, pose accuracy, and computational efficiency compared to prior approaches. Project page: \url{https://linjohnss.github.io/longsplat/}
\end{abstract}

\section{Introduction}
\label{sec:intro}

High-quality 3D reconstruction and novel view synthesis (NVS) are essential for applications such as virtual reality, augmented reality, virtual tourism, and cultural heritage preservation. They also play a crucial role in video editing tasks like stabilization, visual effects, and digital mapping for real estate or pedestrian-level navigation. With the widespread availability of smartphones and action cameras, casually captured videos have emerged as a significant source of 3D content. Unlike professionally acquired datasets, casual videos present challenging characteristics: irregular camera trajectories, long sequences spanning hundreds or thousands of frames, and the absence of reliable camera poses or precise geometric priors.

\begin{figure}[t]
\centering
\resizebox{1.0\columnwidth}{!} 
{
\includegraphics[width=\textwidth]{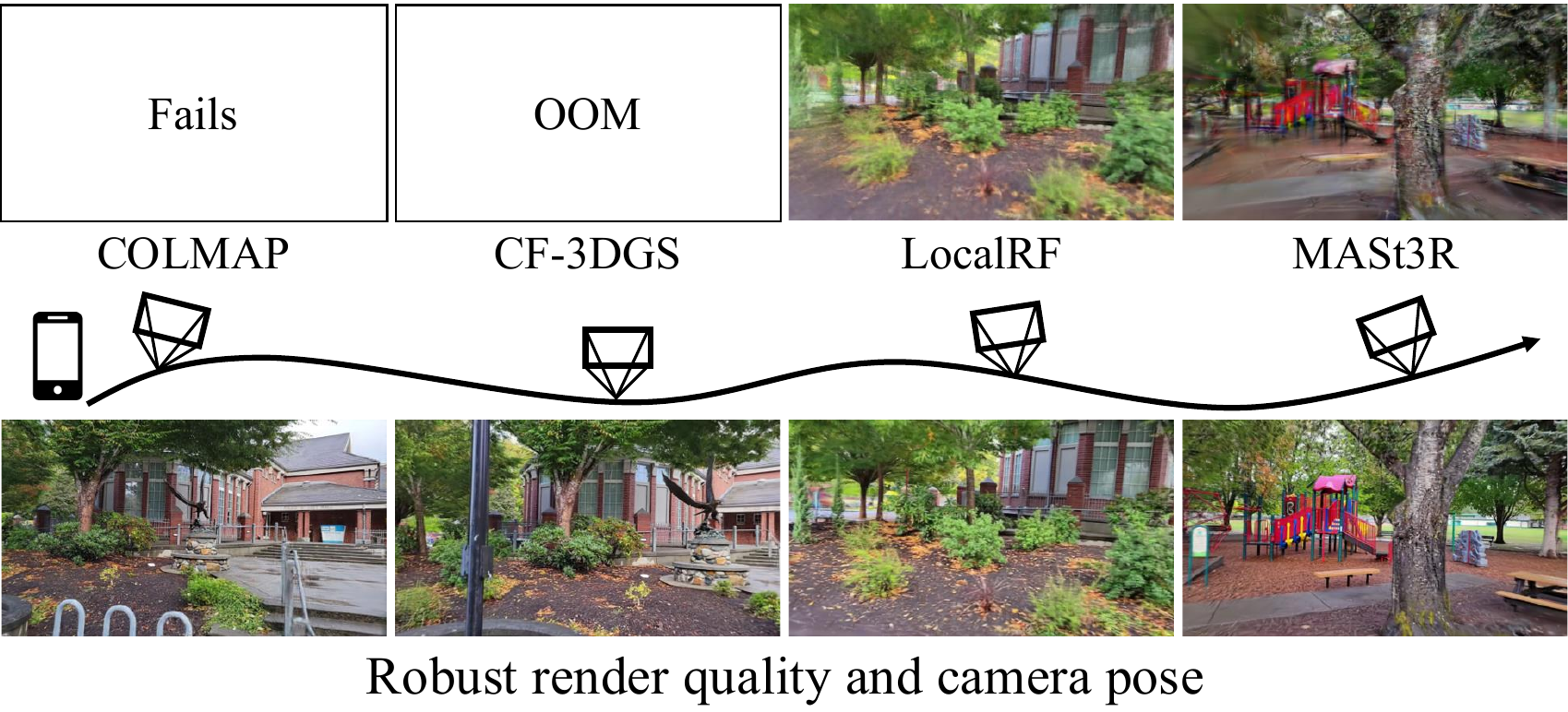}
}
\vspace{-6mm}
\caption{
\textbf{Novel view synthesis for casual long videos.} Existing methods encounter significant challenges when reconstructing scenes from casually captured long videos: COLMAP~\cite{schonberger2016structure} fails due to incorrect camera pose estimation, CF-3DGS~\cite{fu2024colmap} suffers from out-of-memory issues, LocalRF~\cite{meuleman2023progressively} struggles with complex trajectories, and MASt3R~\cite{leroy2024grounding}+Scaffold-GS~\cite{lu2024scaffold} provides inaccurate poses leading to degraded rendering quality. In contrast, LongSplat robustly handles these challenges, yielding accurate camera poses and high-quality novel view synthesis without memory constraints.
}
\label{fig:motivation}
\end{figure}

Addressing novel view synthesis (NVS) from casually captured videos poses two critical challenges: robust camera pose estimation over extended trajectories and efficient representation of large-scale scenes. Traditional methods rely on accurate poses from Structure-from-Motion (SfM) preprocessing, yet as shown in~\cref{fig:motivation}, pipelines like COLMAP~\cite{schonberger2016structure} frequently fail in casual settings. COLMAP-free methods, such as CF-3DGS~\cite{fu2024colmap}, often encounter severe memory constraints, limiting their effectiveness for large-scale scenarios.  Similarly, methods like LocalRF~\cite{meuleman2023progressively} struggle with complex camera trajectories, resulting in fragmented reconstructions. Foundation models like MASt3R~\cite{leroy2024grounding} provide fast initial estimates but suffer inaccuracies and drift in long videos, severely affecting reconstruction quality.

To address these limitations, we introduce \textbf{LongSplat}, a robust unposed 3D Gaussian Splatting (3DGS) \cite{kerbl20233d} framework designed specifically for casual long videos. As illustrated in \cref{teaser}, LongSplat achieves accurate novel view synthesis without relying on provided camera poses. LongSplat departs from traditional pipelines by jointly optimizing camera poses and 3DGS in a unified framework. It integrates a correspondence-guided Pose Estimation Module with 3DGS geometry and photometric refinements to improve pose accuracy, even under large-scale and unstructured camera motion. 
Furthermore, an efficient Octree Anchor Formation mechanism converts dense point clouds into anchors based on spatial density, significantly reducing memory usage while retaining detailed scene structures. These components work together in an incremental joint optimization strategy that avoids local minima and ensures global geometric consistency across extensive sequences.

Extensive experiments on challenging datasets, including Tanks and Temples, Free, and Hike datasets, demonstrate that LongSplat consistently outperforms existing approaches, significantly improving rendering quality and pose accuracy. Compared to conventional methods shown in~\cref{fig:motivation}, LongSplat produces clearer and more coherent reconstructions, effectively addressing pose drift and memory limitations and substantially advancing the state-of-the-art. The
main contributions of our work are:
\begin{itemize} 
\item An incremental joint optimization approach for simultaneous camera pose and 3DGS reconstruction, reducing local minima and ensuring global consistency. 
\item A robust pose estimation module leveraging learned 3D priors for accurate camera pose estimation. 
\item An adaptive Octree Anchor Formation strategy that significantly reduces memory usage while preserving reconstruction quality. 
\end{itemize}

\section{Related Work}
\label{sec:related}

\noindent {\bf Novel View Synthesis.}
Novel View Synthesis (NVS) generates new perspectives from captured images, evolving from early pixel interpolation methods \cite{chen1993view} to depth-based warping techniques \cite{li2013virtual} and 3D reconstruction-based rendering \cite{debevec1996modeling, buehler2001unstructured}. Various representations have been explored, including planes \cite{horry1997tour, hoiem2005automatic}, meshes \cite{hu2020worldsheet, Riegler2020FVS, riegler2021stable}, point clouds \cite{xu2022point, zhang2022differentiable}, and multi-plane images \cite{tucker2020single, zhou2018stereo, li2021mine}.
Neural Radiance Fields (NeRF) \cite{mildenhall2021nerf} revolutionized photorealistic rendering, with subsequent improvements in anti-aliasing \cite{barron2021mip, barron2022mip, barron2023zip, zhang2020nerf++}, reflectance \cite{verbin2022ref, Attal_2023_CVPR}, sparse view training \cite{kim2022infonerf, niemeyer2022regnerf, xu2022sinnerf, yang2023freenerf}, faster training \cite{yu_and_fridovichkeil2021plenoxels, mueller2022instant, reiser2021kilonerf}, and rendering speed \cite{liu2020neural, garbin2021fastnerf, sun2022direct, yu2021plenoctrees}. Recent works have extended NeRF to few-shot scenarios without learned priors \cite{lin2024frugalnerf}, domain-specific applications such as autonomous driving environments \cite{shen2024driveenv}, and dynamic scenes with human pose variations \cite{ma2024humannerf}. 
Point-based methods \cite{xu2022point, zhang2022differentiable, kerbl20233d, luiten2023dynamic}, particularly 3D Gaussian Splatting (3DGS) \cite{kerbl20233d}, enable real-time rendering through explicit representations. Recent advances have extended 3DGS capabilities to dynamic specular scenes with physically-based rendering \cite{fan2025spectromotion}, developed compression techniques for efficient storage and transmission \cite{zhan2025cat}, and improved robustness for unconstrained image scenarios \cite{hou20253d}. However, most approaches still rely on pre-computed camera parameters from SfM \cite{hartley2003multiple, schonberger2016structure, mur2015orb, taketomi2017visual, lin2024frugalnerf}.

\vspace{2pt}
\noindent {\bf Unposed Novel View Synthesis.}
Recent work has aimed to eliminate camera estimation preprocessing. i-NeRF \cite{yen2021inerf} predicts camera poses using pre-trained NeRF. NeRFmm \cite{wang2021nerfmm} jointly optimizes NeRF and camera poses for forward-facing scenes, with SiNeRF \cite{xia2022sinerf} offering improvements. BARF \cite{lin2021barf} and GARF \cite{chng2022garf} address gradient inconsistency through coarse-to-fine positional encoding but require good initialization.
Advanced approaches \cite{bian2023nope, meuleman2023progressively, cheng2023lu, liu2023robust} leverage pre-trained networks for geometric priors, with NoPe-NeRF \cite{bian2023nope} incorporating monocular depth priors and CF-3DGS \cite{fu2024colmap} using progressive optimization. Recent methods have improved robustness in joint optimization of camera poses and scene geometry using decomposed low-rank tensorial representations \cite{chen2024improving} and dynamic radiance fields \cite{liu2023robust}. These methods typically assume small pose perturbations \cite{lin2021barf, chng2022garf}, limited camera motion \cite{xia2022sinerf, wang2021nerfmm}, or additional priors \cite{bian2023nope, meuleman2023progressively, ji2024sfm, cong2025videolifter}, struggling with challenging trajectories, like Free dataset\cite{Kopf2014, ran2024ct}.

\vspace{2pt}
\noindent {\bf Large-scale Novel View Synthesis.}
Extending NVS to large-scale environments introduces memory and computational challenges that NeRF's implicit global representation struggles with. Recent research employs scene partitioning strategies for managing large scenes~\cite{tancik2022block, kerbl2024hierarchical, suzuki2024fed3dgs}. Progressive optimization techniques have been developed for robust view synthesis in large-scale scenes from casually captured videos \cite{meuleman2023progressively}. At the same time, MVS-based approaches have been enhanced to handle generalizable view synthesis at scale \cite{su2024boostmvsnerfs}. For indoor environments, methods like GenRC \cite{li2024genrc} enable room-scale 3D reconstruction from sparse image collections. 3DGS offers explicit representation advantages through Gaussian primitive. VastGaussian \cite{lin2024vastgaussian} divides scenes into separately optimized blocks\cite{kerbl20233d}. Scaffold-GS \cite{lu2024scaffold} introduces anchor-based Gaussian representation with fixed-resolution grids, though it requires SfM initialization. Octree-GS \cite{ren2024octree} implements fixed-level octrees with preset resolutions but similarly depends on SfM. Unlike these approaches, our method dynamically adjusts voxel size based on point cloud density, without dependency on SfM, and addresses unposed, large-scale, casually captured videos through adaptive Octree Anchor Formation.

\vspace{2pt}
\noindent {\bf Casual Long Videos.}
Casual long videos present unique challenges: free-moving trajectories, lack of pose information, and continuously expanding scenes. LocalRF \cite{meuleman2023progressively} addresses these through progressive localized field construction but suffers from slow training and fragmentation under irregular camera movements.
3D Foundation Models~\cite{wang2024spann3r}, including DUSt3R \cite{wang2024dust3r}, MASt3R \cite{leroy2024grounding}, Fast3r \cite{yang2025fast3r}, and CUT3R \cite{wang2025continuous}, estimate poses and geometry directly but accumulate errors in long sequences. LongSplat treats foundation model outputs as soft priors, jointly optimizing them with 3D Gaussian Splatting while progressively correcting poses and geometry through combined PnP and optimization strategies.

\begin{figure*}[t]
\centering
\vspace{-6mm}
\includegraphics[width=\textwidth]{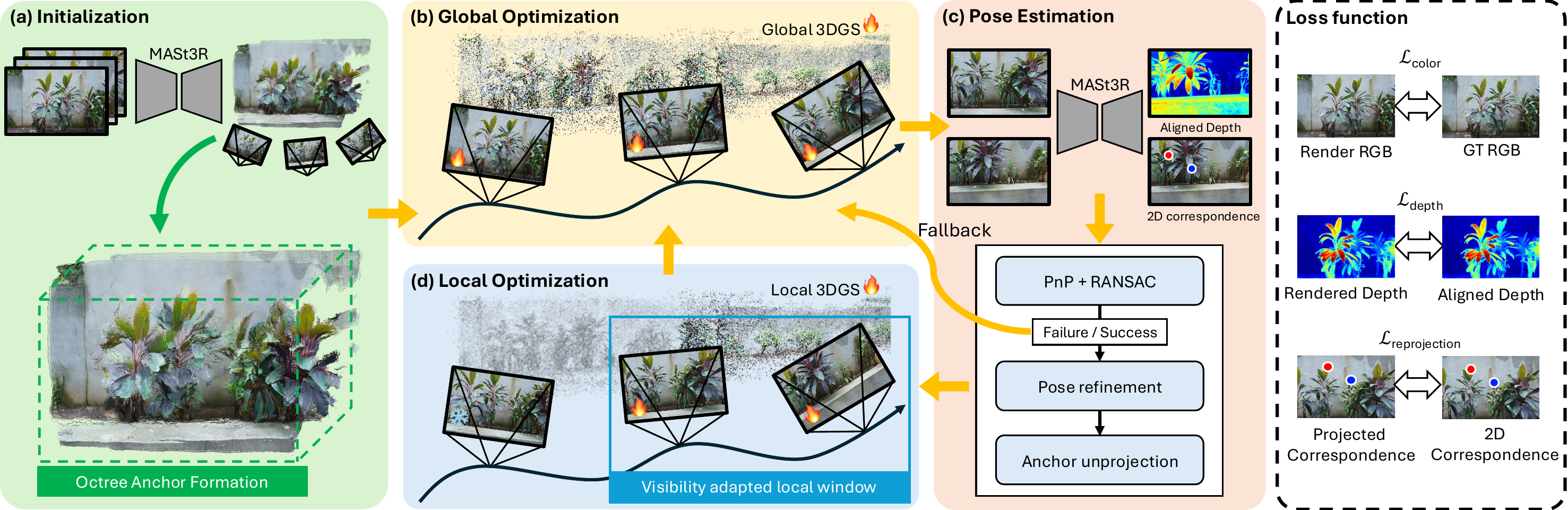}
\vspace{-6mm}
\caption{\textbf{Overview of the LongSplat framework.} 
Given a casually captured long video without known poses, LongSplat incrementally reconstructs the scene through tightly coupled pose estimation and 3D Gaussian Splatting. 
(a) Initialization converts MASt3R~\cite{leroy2024grounding} global aligned point cloud into an octree-anchored 3DGS. 
(b) Global Optimization jointly refines all camera poses and 3D Gaussians for global consistency.
(c) Pose estimation estimates each new frame pose via correspondence-guided PnP, applies photometric refinement, and updates octree anchors using unprojected points. If PnP fails, a fallback triggers global re-optimization to recover.
(d) Incremental Optimization alternates between Local Optimization within a visibility-adapted window and periodic Global Optimization to propagate consistent updates across frames.
(e) All optimization stages leverage a unified objective composed of photometric loss, depth loss, and reprojection loss to ensure accurate geometry and appearance reconstruction.
}
\label{fig:framework}
\end{figure*}

\section{Method}
\label{sec:method}
\vspace{-1pt}

LongSplat reconstructs long video sequences with unknown camera poses and unconstrained trajectories through a fully incremental pipeline based on octree-anchored 3D Gaussian Splatting. The process begins with octree anchor formation, where per-frame dense point clouds are structured into an adaptive representation. Next, camera poses are estimated and refined using correspondence-guided initialization and photometric alignment. Finally, the reconstruction alternates between local optimization, which updates Gaussians within a visibility-adapted window, and global refinement, which ensures long-term consistency. This design allows LongSplat to robustly handle long, unconstrained trajectories while adapting to scene complexity and minimizing drift.


\vspace{-1pt}
\subsection{Preliminaries}
\vspace{-1pt}
\noindent {\bf Gaussian Splatting.}
3D Gaussian Splatting (3DGS)~\cite{kerbl20233d} represents the scene as a set of 3D Gaussians, each defined by a center $\mu \in \mathbb{R}^3$, a covariance matrix $\Sigma$, color, scale, rotation, and opacity. The covariance is factorized into a rotation $R \in SO(3)$ and a diagonal scale matrix $S$, giving:
\vspace{-1pt}
\small
\begin{equation}
G(x) = e^{-\frac{1}{2}(x - \mu)^\top \Sigma^{-1} (x - \mu)}, \quad \Sigma = R S S^\top R^\top.
\vspace{-1pt}
\end{equation}
\normalsize
This parameterization allows each Gaussian to adaptively capture local scene geometry.

To render the scene, each Gaussian is projected onto the image plane using the camera pose $W$, resulting in a 2D Gaussian with covariance $\Sigma_\text{2D} = J W \Sigma W^\top J^\top$,
where $J$ is the Jacobian of the projective transformation. The final rendered color and depth are computed via alpha blending:
\vspace{-1pt}
\small
\begin{equation}
\begin{aligned}
C = \sum_{i=1}^{N} c_i \,\alpha_i \prod_{j=1}^{i-1} \bigl(1-\alpha_j\bigr), \quad 
D = \sum_{i=1}^{N} d_i \,\alpha_i \prod_{j=1}^{i-1} \bigl(1-\alpha_j\bigr),
\vspace{-1pt}
\end{aligned}
\label{eq:rendering}
\end{equation}
\normalsize
where $c_i$ and $\alpha_i$ denote the color and opacity of the $i$-th Gaussian, respectively. 
$d_i$ denotes the depth value along the ray at the Gaussian's center.

\vspace{2pt}
\noindent {\bf Anchor-based 3D Gaussian Splatting.}
To enhance memory efficiency and robustness in large scenes, Scaffold-GS \cite{lu2024scaffold} introduces the anchor-based 3DGS representation. Instead of directly maintaining individual Gaussians, the scene is first divided into sparse voxels, each acting as an anchor. From each anchor, $k$ Gaussians are initialized with positions relative to the anchor center:
\vspace{-1pt}
\small
\begin{equation}
\{\mu_0, \mu_1, \dots, \mu_{k-1}\} = x_v + \{O_0, O_1, \dots, O_{k-1}\} \cdot l_v,
\vspace{-1pt}
\end{equation}
\normalsize
where $x_v$ denotes the anchor position, $\{O_i\}$ are offset vectors, and $l_v$ is a scaling factor. Each Gaussian’s opacity, rotation, scale, and color are decoded from an anchor feature through lightweight MLPs. For opacity, the formulation is:
\vspace{-1pt}
\small
\begin{equation}
\{\alpha_0, \alpha_1, \dots, \alpha_{k-1}\} = F_\alpha(\hat{f}_v, \Delta v_c, \hat{d}_v),
\vspace{-1pt}
\end{equation}
\normalsize
where $F_\alpha$ is an MLP taking the anchor feature $\hat{f}_v$, the relative view distance $\Delta v_c$, and the view direction $\hat{d}_v$ as inputs.

\vspace{2pt}
\noindent {\bf Anchor Initialization.}
In traditional Scaffold-GS, initial anchors are derived from sparse SfM point clouds. Points are voxelized to form anchor centers:
\vspace{-1pt}
\small
\begin{equation}
V = \{v \mid v = \lfloor \frac{p}{\epsilon} \rfloor \cdot \epsilon, \forall p \in P\},
\vspace{-1pt}
\end{equation}
\normalsize
where $P$ is the SfM point cloud and $\epsilon$ is the voxel size. Each anchor holds a local feature, managing its associated Gaussians. This design ensures structured densification and pruning, adapting Gaussian density to scene complexity and improving both memory and rendering efficiency.

\begin{figure}[t]
\centering
\vspace{-6mm}
\resizebox{1.0\columnwidth}{!} 
{
\includegraphics[width=\textwidth]{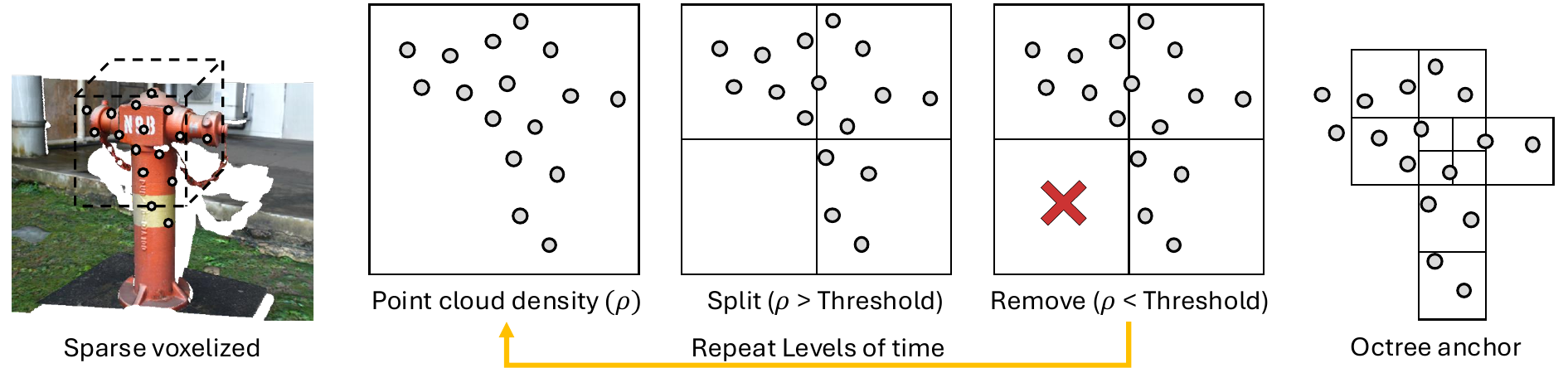}
}
\vspace{-6mm}
\caption{\textbf{Visualization of our proposed Octree Anchor Formation strategy.}
Given an initial sparse voxelized point cloud, we iteratively perform density-guided adaptive voxel splitting and pruning. Voxels with point cloud density ($\rho$) exceeding a threshold are split, while those with density below the threshold are pruned. Repeated across multiple octree levels, this adaptive octree anchor design significantly reduces memory usage, allowing efficient representation and rendering of large-scale scenes.
}
\label{fig:octree_anchor}
\end{figure}

\subsection{Octree Anchor Formation}
\label{sec:octree_anchor}

In large-scale casual video settings, memory efficiency and scene adaptability are essential. Our Octree Anchor Formation dynamically adjusts spatial resolution based on observed geometry, enabling scalable and redundant-free anchor management. LongSplat constructs structured anchors from MASt3R's per-frame dense point clouds using an adaptive octree (\cref{fig:framework} (a)). Unlike Scaffold-GS, which relies on a fixed-resolution grid, we progressively subdivide space based on local point density. Each point cloud $\mathbf{P} = \{p_i\}$ is voxelized into a sparse grid at resolution $\epsilon_0$. Voxels exceeding a density threshold $\tau_\text{split}$ split into 8 smaller voxels:
\vspace{-1pt}
\small
\begin{equation}
\epsilon_{l+1} = \frac{1}{2}\epsilon_l.
\vspace{-1pt}
\end{equation}
\normalsize
This process repeats up to a maximum level $L$. Low-density voxels (density $\rho_v < \tau_\text{prune}$) are removed to reduce redundancy (\cref{fig:octree_anchor}).

Each anchor inherits a spatial scale proportional to its voxel size, ensuring coarse anchors for sparsely observed areas and finer anchors for detailed regions:
\vspace{-1pt}
\small
\begin{equation}
s_v \propto \epsilon_v.
\vspace{-1pt}
\end{equation}
\normalsize

To further prevent unnecessary duplication, newly generated anchors are compared to existing ones. If significant spatial overlap exists, the new anchor is discarded. This density-adaptive, duplication-free octree formation ensures compact memory usage while preserving adaptive resolution across scenes.

\begin{figure}[t]
\centering
\vspace{-6mm}
\resizebox{1.0\columnwidth}{!} 
{
\includegraphics[width=\textwidth]{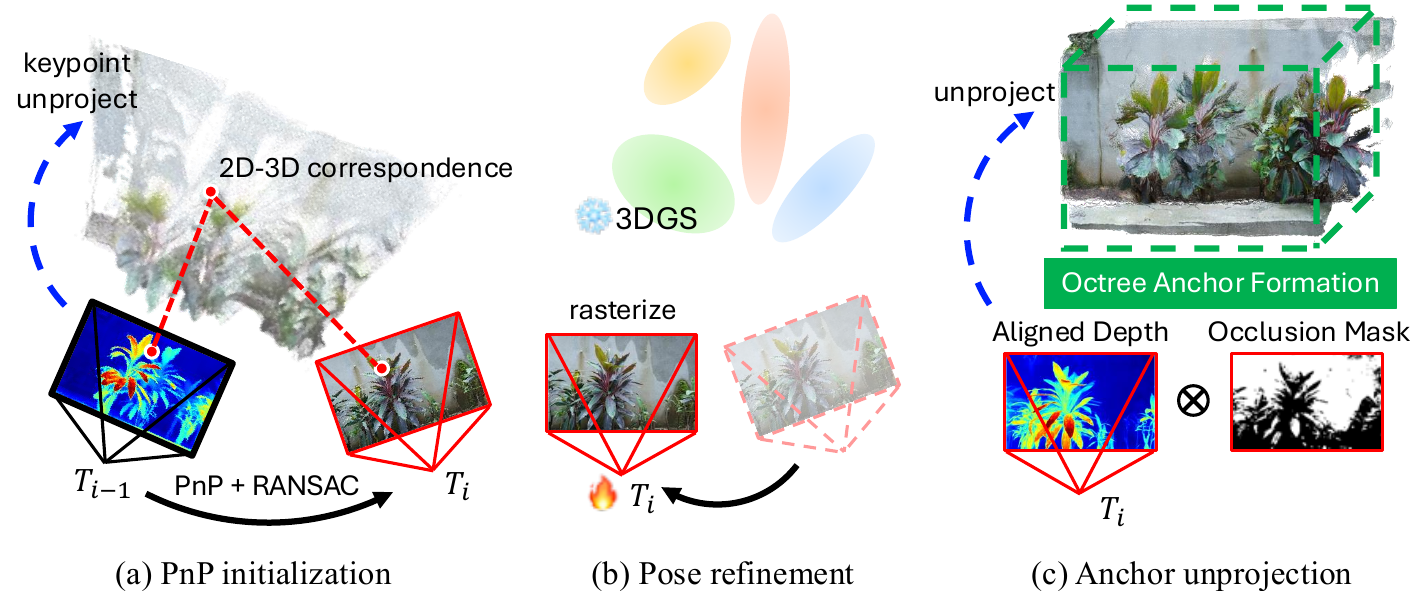}
}
\vspace{-6mm}
\caption{
\textbf{Detailed illustration of our camera pose estimation.}
(a) PnP initialization: Given correspondences between the predicted 3D anchor points from frame $T_{i-1}$ and the 2D keypoints detected in frame $T_i$ we employ PnP with RANSAC to robustly estimate an initial camera pose. (b) Pose refinement: The estimated pose is further refined by rasterizing the 3DGS scene and iteratively minimizing reprojection error to enhance pose accuracy. (c) Anchor unprojection: Newly observed regions are detected via an occlusion mask, computed by forward-warping the previous frame's rendered depth. These regions are unprojected into 3D and converted into anchors via Octree Anchor Formation.
}
\label{fig:pose_estimation}
\end{figure}

\subsection{Pose Estimation module}
\label{sec:pose_estimation}
Accurate and robust camera pose estimation is essential for consistent reconstruction in unposed long video settings. 
We estimate each pose using 2D-3D correspondences derived from MASt3R, followed by photometric refinement against the current 3D Gaussian scene to maintain coherence across evolving 3D structures (\cref{fig:framework} (c)).

For each new frame $t$, MASt3R provides 2D correspondences $\{(x_i, x_i')\}$ between frame $t$ and $t-1$, allowing back-projection of matched points $x_i$ to 3D via:
\vspace{-1pt}
\small
\begin{equation}
X_i = D_{t-1}(x_i) \cdot K^{-1} \tilde{x}_i.
\vspace{-1pt}
\end{equation}
\normalsize
These 2D-3D correspondences $\{(x_i', X_i)\}$ are used to solve the initial pose $T_t$ via PnP (\cref{fig:pose_estimation} (a)), followed by photometric refinement that minimizes (\cref{fig:pose_estimation} (b)):
\vspace{-1pt}
\small
\begin{equation}
\mathcal{L}_\text{photo} = \sum_{p \in \Omega} \|I_t(p) - \hat{I}_t(p)\|^2,
\vspace{-1pt}
\end{equation}
\normalsize
where $I_t$ is the observed frame and $\hat{I}_t$ is the rendering using the current 3DGS. This ensures the pose aligns with the evolving scene.

To correct MASt3R’s depth scale drift, we compute a scale factor $\hat{s}_t$ by comparing the rendered depth $D_{t-1}$ and MASt3R’s aligned depth $D_t^\text{align}$:
\vspace{-1pt}
\small
\begin{equation}
\hat{s}_t = \frac{\langle D_{t-1}, D_t^\text{align} \rangle}{\langle D_t^\text{align}, D_t^\text{align} \rangle}.
\vspace{-1pt}
\end{equation}
\normalsize
This rescaled depth ensures consistent scale across frames.

As the camera moves, newly visible regions are detected via an occlusion mask $M_\text{occ}$, derived by forward-warping $D_{t-1}$ to frame $t$ and comparing it to the rescaled depth $D^{\text{MASt3R}}_t$ (\cref{fig:pose_estimation} (c)). Newly visible pixels are unprojected into 3D using:
\vspace{-1pt}
\small
\begin{equation}
p_i = D^{\text{MASt3R}}_{t,\mathbf{u}_i} \cdot \mathbf{K}^{-1}\mathbf{u}_i.
\vspace{-1pt}
\end{equation}
\normalsize
These new points are converted into octree anchors using the \emph{Octree Anchor Formation} described in Sec.~\ref{sec:octree_anchor}, with overlapping anchors removed to avoid redundancy (\cref{fig:pose_estimation} (c)). This process incrementally expands the scene while maintaining structural regularity.

\begin{figure}[t]
\centering
\resizebox{1.0\columnwidth}{!} 
{
\includegraphics[width=\textwidth]{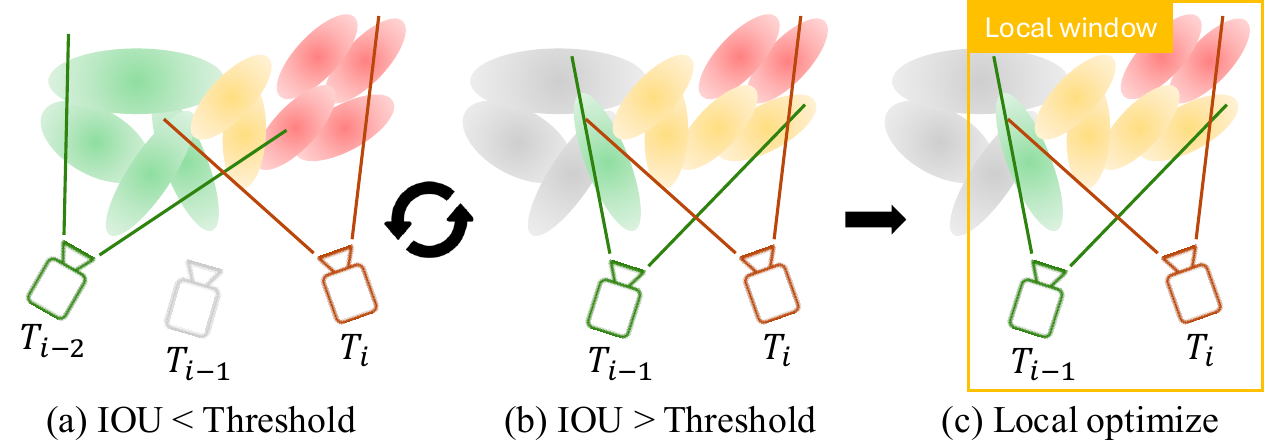}
}
\vspace{-6mm}
\caption{\textbf{Illustration of our Visibility-Adapted Local Window strategy for local optimization.}
To ensure balanced training of the 3D Gaussians, we dynamically define the optimization window based on anchor visibility overlap. Specifically, we compute the Intersection-over-Union (IoU) of visible anchors between consecutive views. Suppose the visibility IoU is below a certain threshold (a). In that case, the local optimization window is adjusted by removing the earliest frame, iteratively repeating until a suitable window with IoU above the threshold is found (b). This approach ensures balanced training coverage and enhances local reconstruction details during optimization (c).
}
\label{fig:visibility_local_window}
\end{figure}

\subsection{Incremental Joint Optimization}

To handle casually captured long videos, LongSplat adopts a progressive incremental optimization framework that alternates between per-frame local reconstruction and cross-frame global consistency refinement.

\vspace{2pt}
\noindent {\bf Initialization.}
We begin with a small set of initial frames. Camera poses and dense point clouds for these frames are estimated using MASt3R~\cite{leroy2024grounding}, followed by converting the point cloud into an initial octree-anchored 3DGS using the proposed Octree Anchor Formation (\cref{fig:framework} (a). When camera intrinsics are unavailable, we directly adopt MASt3R’s estimated focal length.

\vspace{2pt}
\noindent {\bf Global Optimization.}
After initialization, we jointly optimize all 3D Gaussian parameters and camera poses across all processed frames (\cref{fig:framework} (b)). This global optimization ensures geometric consistency across the entire sequence, reducing accumulated pose drift and local misalignments.

\vspace{2pt}
\noindent {\bf Frame Insertion and Pose Estimation.}
As new frames arrive, we estimate their poses using the correspondence-guided PnP initialization and refinement strategy described in Sec.~\ref{sec:pose_estimation}. If PnP fails due to insufficient feature correspondences or poor initialization, we trigger a fallback mechanism that re-optimizes all past frames globally before retrying pose estimation. This iterative fallback enhances robustness under challenging motion or weak texture (\cref{fig:framework} (c)).

\vspace{2pt}
\noindent {\bf Local Optimization with Visibility-Adaptive Window.}
Once the pose is estimated, we optimize only the Gaussians visible in the new frame’s frustum, while constraining them with observations from nearby frames in a dynamically selected \emph{visibility-adapted local window} (\cref{fig:visibility_local_window}). Covisibility between frames is measured by:
\vspace{-1pt}
\small
\begin{equation}
\text{IoU}(t, t') = \frac{| \mathcal{V}(t) \cap \mathcal{V}(t') |}{| \mathcal{V}(t) \cup \mathcal{V}(t') |},
\vspace{-1pt}
\end{equation}
\normalsize
where $\mathcal{V}(t)$ denotes the set of Gaussians visible in frame $t$. Frames with covisibility below a threshold $\tau$ are excluded from the window. This adaptive mechanism ensures local Gaussians are consistently supervised by reliable multi-view constraints, balancing efficiency and accuracy.

\vspace{2pt}
\noindent {\bf Final Global Refinement.} In the final step, a final global refinement jointly optimizes all Gaussians and camera poses over the sequence. This final pass further improves both rendering quality and long-range pose consistency.

\vspace{2pt}
\noindent {\bf Depth and Reprojection Losses.}
To provide additional supervision in newly revealed regions, where multi-view observations are insufficient, we introduce two regularization terms. A monocular depth loss encourages rendered depth to match MASt3R’s scale-aligned depth prior:
\vspace{-1pt}
\small
\begin{equation}
\mathcal{L}_\text{depth} = \| D^\text{rendered} - D^\text{MASt3R} \|^2.
\vspace{-1pt}
\end{equation}
\normalsize
Additionally, a keypoint reprojection loss enforces alignment between projected 3D keypoints and their 2D observations:
\vspace{-1pt}
\small
\begin{equation}
\mathcal{L}_\text{reprojection} = \sum_{k} \| \pi(\mathbf{X}_k) - \mathbf{u}_k \|^2,
\vspace{-1pt}
\end{equation}
\normalsize
where $\pi(\cdot)$ denotes projection using the current pose.

\vspace{2pt}
\noindent {\bf Total Loss.}
Throughout the entire incremental reconstruction pipeline, each processed frame is optimized using the following objective:
\vspace{-1pt}
\small
\begin{equation}
\mathcal{L}_\text{total} = \mathcal{L}_\text{photo} + \lambda_\text{depth}\mathcal{L}_\text{depth} + \lambda_\text{reprojection}\mathcal{L}_\text{reprojection},
\vspace{-1pt}
\end{equation}
\normalsize
This combined loss applies to both local and global optimization stages, ensuring coherent multi-view, robust pose refinement, and stable geometry reconstruction across the evolving scene.


\section{Experiments}
\label{sec:experiments}

\subsection{Experimental Setup}

\begin{table*}[t]
\vspace{-6mm}
\caption{\textbf{Quantitative comparison on the Free dataset~\cite{wang2023f2} across various baseline methods.} Methods such as CF-3DGS~\cite{fu2024colmap} frequently encounter out-of-memory issues, denoted by ``-''. Our method consistently outperforms all baselines across diverse scenes, delivering superior rendering quality and robustness, especially in challenging environments characterized by complex camera trajectories and varied geometric structures. ``*'': Initialized with MASt3R poses, then jointly optimized.}
\label{tab:free_quantitative}
\centering
\vspace{-3mm}
\resizebox{\textwidth}{!}{%
\setlength{\tabcolsep}{2pt}
\begin{tabular}{l|ccc ccc ccc ccc|ccc ccc ccc ccc}
\toprule
\multirow{3}{*}{Scenes} & \multicolumn{3}{c}{COLMAP~\cite{schonberger2016structure}} & \multicolumn{3}{c}{COLMAP~\cite{schonberger2016structure}} & \multicolumn{3}{c}{MASt3R~\cite{leroy2024grounding}} & \multicolumn{3}{c|}{MASt3R~\cite{leroy2024grounding}} & \multicolumn{3}{c}{\multirow{2}{*}{CF-3DGS~\cite{fu2024colmap}}} & \multicolumn{3}{c}{\multirow{2}{*}{NoPe-NeRF~\cite{bian2023nope}}} & \multicolumn{3}{c}{\multirow{2}{*}{LocalRF~\cite{meuleman2023progressively}}} & \multicolumn{3}{c}{\multirow{2}{*}{Ours}} \\
& \multicolumn{3}{c}{+ F2-NeRF~\cite{wang2023f2}} & \multicolumn{3}{c}{+ Scaffold-GS~\cite{lu2024scaffold}} & \multicolumn{3}{c}{+ Scaffold-GS~\cite{lu2024scaffold}} & \multicolumn{3}{c|}{+ Scaffold-GS~\cite{lu2024scaffold}*} & & & & & & & & & & & & \\
\cmidrule(lr){2-4} \cmidrule(lr){5-7} \cmidrule(lr){8-10} \cmidrule(lr){11-13} \cmidrule(lr){14-16} \cmidrule(lr){17-19} \cmidrule(lr){20-22} \cmidrule(lr){23-25}
& PSNR$\uparrow$ & SSIM$\uparrow$ & LPIPS$\downarrow$ & PSNR$\uparrow$ & SSIM$\uparrow$ & LPIPS$\downarrow$ & PSNR$\uparrow$ & SSIM$\uparrow$ & LPIPS$\downarrow$ & PSNR$\uparrow$ & SSIM$\uparrow$ & LPIPS$\downarrow$ & PSNR$\uparrow$ & SSIM$\uparrow$ & LPIPS$\downarrow$ & PSNR$\uparrow$ & SSIM$\uparrow$ & LPIPS$\downarrow$ & PSNR$\uparrow$ & SSIM$\uparrow$ & LPIPS$\downarrow$ & PSNR$\uparrow$ & SSIM$\uparrow$ & LPIPS$\downarrow$ \\
\midrule
Grass   & 23.44 & 0.58 & 0.45 & 26.75 & 0.82 & 0.20 & 22.65 & 0.61 & 0.34 & 25.06 & 0.79 & 0.21 & -     & -    & -    & 16.39 & 0.27 & 0.81 & 18.84 & 0.35 & 0.60 & \textbf{26.16} & \textbf{0.80} & \textbf{0.22} \\
Hydrant & 23.75 & 0.74 & 0.28 & 26.66 & 0.86 & 0.12 & 23.22 & 0.71 & 0.21 & 25.68 & 0.83 & 0.12 & -     & -    & -    & 17.94 & 0.43 & 0.66 & 19.19 & 0.48 & 0.48 & \textbf{24.69} & \textbf{0.79} & \textbf{0.18} \\
Lab     & 24.34 & 0.83 & 0.26 & 28.27 & 0.92 & 0.10 & 20.66 & 0.74 & 0.25 & 22.42 & 0.80 & 0.18 & -     & -    & -    & 17.42 & 0.52 & 0.63 & 17.22 & 0.55 & 0.47 & \textbf{27.11} & \textbf{0.87} & \textbf{0.15} \\
Pillar  & 28.05 & 0.79 & 0.23 & 31.75 & 0.90 & 0.12 & 23.95 & 0.70 & 0.28 & 22.88 & 0.67 & 0.24 & 14.55  & 0.40 & 0.66 & 18.88 & 0.44 & 0.75 & 22.98 & 0.59 & 0.49 & \textbf{30.44} & \textbf{0.88} & \textbf{0.16} \\
Road    & 26.03 & 0.80 & 0.27 & 30.45 & 0.92 & 0.10 & 24.23 & 0.73 & 0.25 & 25.05 & 0.78 & 0.27 & -     & -    & -    & 17.48 & 0.44 & 0.79 & 20.68 & 0.54 & 0.56 & \textbf{27.73} & \textbf{0.84} & \textbf{0.20} \\
Sky     & 25.10 & 0.86 & 0.24 & 28.34 & 0.92 & 0.12 & 23.26 & 0.80 & 0.22 & 25.37 & 0.88 & 0.14 & - & - & - & 16.18 & 0.51 & 0.65 & 18.76 & 0.60 & 0.46 & \textbf{28.07} & \textbf{0.91} & \textbf{0.13} \\
Stair   & 28.14 & 0.84 & 0.22 & 32.13 & 0.93 & 0.10 & 23.35 & 0.71 & 0.30 & 24.46 & 0.79 & 0.28 & 13.41 & 0.41 & 0.63 & 19.14 & 0.47 & 0.69 & 23.55 & 0.66 & 0.38 & \textbf{31.00} & \textbf{0.89} & \textbf{0.16} \\
\midrule
Avg.     & 25.55 & 0.78 & 0.28 & 29.19 & 0.90 & 0.12 & 23.05 & 0.72 & 0.27 & 24.42 & 0.79 & 0.21 & 13.98 & 0.41 & 0.65 & 17.63 & 0.44 & 0.71 & 20.17 & 0.54 & 0.49 & \textbf{27.88} & \textbf{0.85} & \textbf{0.17} \\
\bottomrule
\end{tabular}
}
\end{table*}

\begin{figure*}[t]
\centering
\resizebox{1.0\textwidth}{!} 
{
\includegraphics[width=\textwidth]{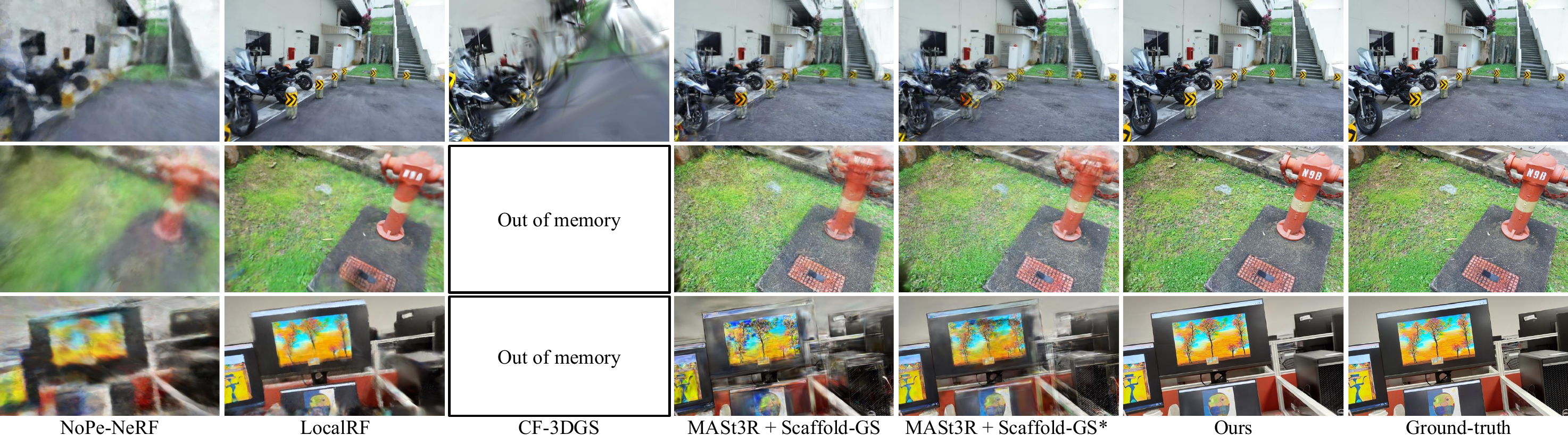}
}
\vspace{-7mm}
\caption{\textbf{Qualitative comparison on the Free dataset~\cite{wang2023f2}.} We compare our method with state-of-the-art approaches including NoPe-NeRF~\cite{bian2023nope}, LocalRF~\cite{meuleman2023progressively}, CF-3DGS~\cite{fu2024colmap}, and MASt3R~\cite{leroy2024grounding} combined with Scaffold-GS~\cite{lu2024scaffold}. CF-3DGS fails due to memory constraints (OOM), and other baseline methods exhibit artifacts or blurry reconstructions. In contrast, our method produces results closest to the ground truth, demonstrating clearer details, accurate geometry, and visually consistent rendering, particularly under challenging scene structures and complex camera trajectories. ``*'': Initialized with MASt3R poses, then jointly optimized.}
\label{fig:free_visual}
\end{figure*}

\noindent {\bf Datasets.}
We evaluate LongSplat on three challenging real-world datasets with varying difficulty levels:
\begin{itemize} 
\item \textbf{Tanks and Temples~\cite{knapitsch2017tanks} (Standard)}: Eight scenes with smooth, forward-facing camera trajectories, evaluated at full resolution. Every 8$^{\text{th}}$ frame is used for testing.
\item \textbf{Free Dataset~\cite{wang2023f2} (Moderate)}: Seven handheld videos featuring complex, unconstrained trajectories with multiple foreground objects, evaluated at 1/2 resolution. Frequent scene changes make memory-efficient 3D representation essential. Every 8$^{\text{th}}$ frame is tested.
\item \textbf{Hike Dataset~\cite{meuleman2023progressively} (Hard)}: Long videos with hundreds to thousands of frames, complex trajectories, and detailed geometry, evaluated at 1/4 resolution. The scale and duration demand adaptive memory management. Every 10$^{\text{th}}$ frame is used for testing.
\end{itemize}

\vspace{2pt}
\noindent {\bf Evaluation Metrics.} We evaluate novel view synthesis quality using PSNR, SSIM~\cite{wang2004image}, and LPIPS~\cite{zhang2018unreasonable}. Pose accuracy is measured with Absolute Trajectory Error (ATE) and Relative Pose Error (RPE), using COLMAP poses as ground truth. We also report model size, training time, and FPS to assess computational efficiency.


\vspace{2pt}
\noindent {\bf Baselines.} We compare LongSplat with COLMAP-based methods (COLMAP~\cite{schonberger2016structure}+F2-NeRF~\cite{wang2023f2} / 3DGS~\cite{kerbl20233d} / Scaffold-GS~\cite{lu2024scaffold}) and unposed methods (NoPe-NeRF~\cite{bian2023nope}, LocalRF~\cite{meuleman2023progressively}, CF-3DGS~\cite{fu2024colmap}). Additionally, we evaluate a naïve baseline combining MASt3R's~\cite{leroy2024grounding} predicted point cloud and poses with Scaffold-GS. During training, camera poses are either fixed (MASt3R + Scaffold-GS) or jointly optimized (MASt3R + Scaffold-GS*).

\begin{table}[t]
\centering
\small
\renewcommand{\arraystretch}{0.8}
\caption{\textbf{Quantitative evaluation of camera pose estimation accuracy on the Free dataset~\cite{wang2023f2}.} Our method achieves superior performance across most scenes, significantly reducing pose errors compared to state-of-the-art approaches. ``*'': Initialized with MASt3R poses, then jointly optimized.}
\vspace{-3mm}
\label{tab:free_pose_quantitative}
\footnotesize
\begin{tabular}{l|ccc}
\toprule
Method & RPE$_t$$\downarrow$ & RPE$_r$$\downarrow$ & ATE$\downarrow$ \\
\midrule
MASt3R~\cite{leroy2024grounding} + Scaffold-GS~\cite{lu2024scaffold}   & 0.162 & 0.265 & 0.013 \\
MASt3R~\cite{leroy2024grounding} + Scaffold-GS~\cite{lu2024scaffold}*  & 0.083 & 0.176 & 0.008 \\
CF-3DGS~\cite{fu2024colmap}     & 0.234 & 3.442 & 0.022 \\
NoPe-NeRF~\cite{bian2023nope}     & 6.231 & 4.822 & 0.576 \\
LocalRF~\cite{meuleman2023progressively}               & 0.754 & 7.086 & 0.035 \\
Ours                           & \textbf{0.028} & \textbf{0.103} & \textbf{0.004} \\
\bottomrule
\end{tabular}%
\end{table}

\begin{figure}[t]
\centering
\resizebox{1.0\columnwidth}{!} 
{
\includegraphics[width=\textwidth]{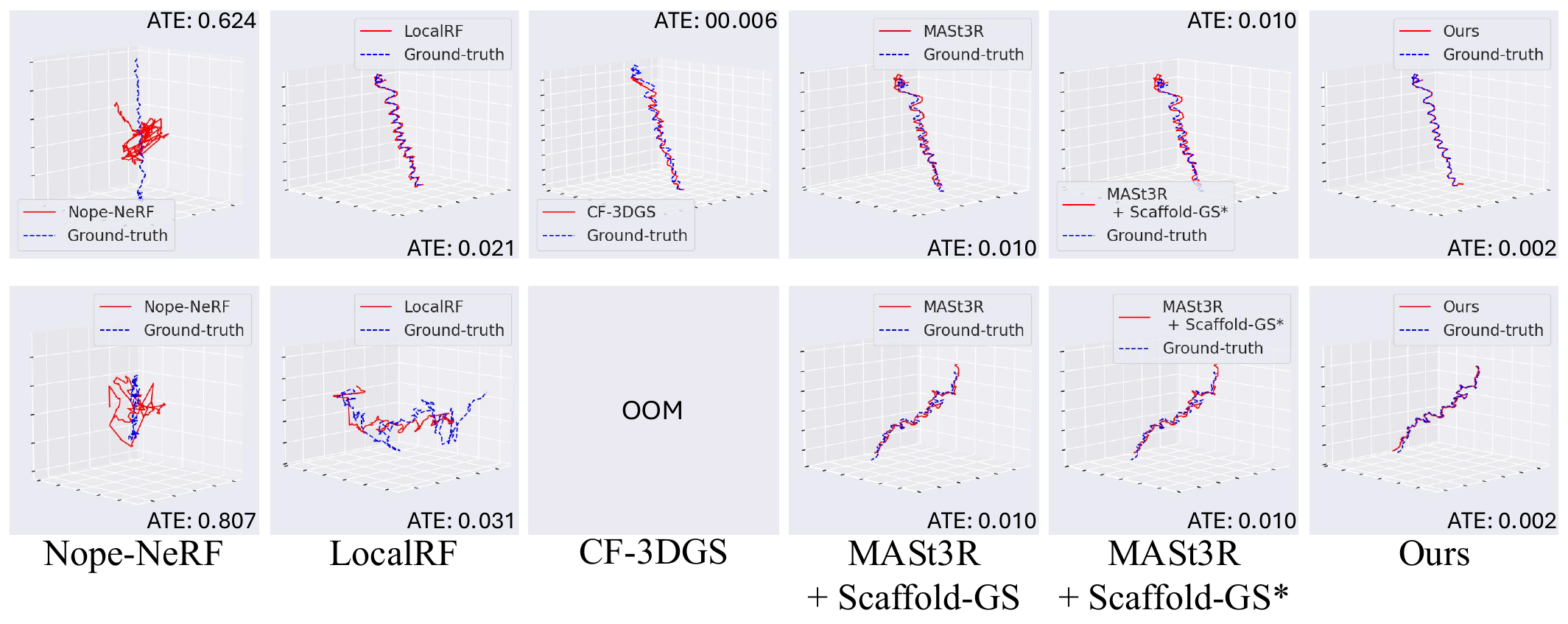}
}
\vspace{-8mm}
\caption{
\textbf{Visualization of camera trajectories on Free dataset~\cite{wang2023f2}.} CF-3DGS~\cite{fu2024colmap} encounters OOM and fails for long sequences, whereas our method reliably estimates accurate, stable trajectories, demonstrating superior robustness.
}
\label{fig:free_traj_visual}
\end{figure}


\begin{table}[t]
\centering
    \small
\vspace{-3mm}
    \renewcommand{\arraystretch}{0.8}

\caption{\textbf{Quantitative evaluation of novel view synthesis quality on the Tanks and Temples dataset~\cite{knapitsch2017tanks}.} Our proposed LongSplat consistently surpasses existing methods across multiple scenes.}
\vspace{-3mm}
\label{tab:tnt_average_quantitative}
\footnotesize
\resizebox{\columnwidth}{!}{%
\begin{tabular}{l|cccccc}
\toprule
Method & PSNR$\uparrow$ & SSIM$\uparrow$ & LPIPS$\downarrow$ & RPE$_t$$\downarrow$ & RPE$_r$$\downarrow$ & ATE$\downarrow$ \\
\midrule
COLMAP+3DGS~\cite{kerbl20233d} & 30.21 & 0.92 & 0.10 & --   & --   & --    \\
MASt3R~\cite{leroy2024grounding} + Scaffold-GS~\cite{lu2024scaffold} & 28.67 & 0.79 & 0.21 & 0.166 & 0.168 & 0.006  \\
MASt3R~\cite{leroy2024grounding} + Scaffold-GS~\cite{lu2024scaffold}* & 30.92 & 0.90 & 0.13 & 0.047 & 0.103 & 0.005 \\
NoPe-NeRF~\cite{bian2023nope}   & 26.34 & 0.74 & 0.39 & 0.080 & \textbf{0.038} & 0.006 \\
CF-3DGS~\cite{fu2024colmap}     & 31.28 & 0.93 & 0.09 & 0.041 & 0.069 & 0.004 \\
Ours        & \textbf{32.83} & \textbf{0.94} & \textbf{0.08} & 0\textbf{0.032} & 0.068 & \textbf{0.003} \\
\bottomrule
\end{tabular}%
}
\end{table}

\begin{figure}
\centering
\resizebox{1.0\columnwidth}{!} 
{
\includegraphics[width=\textwidth]{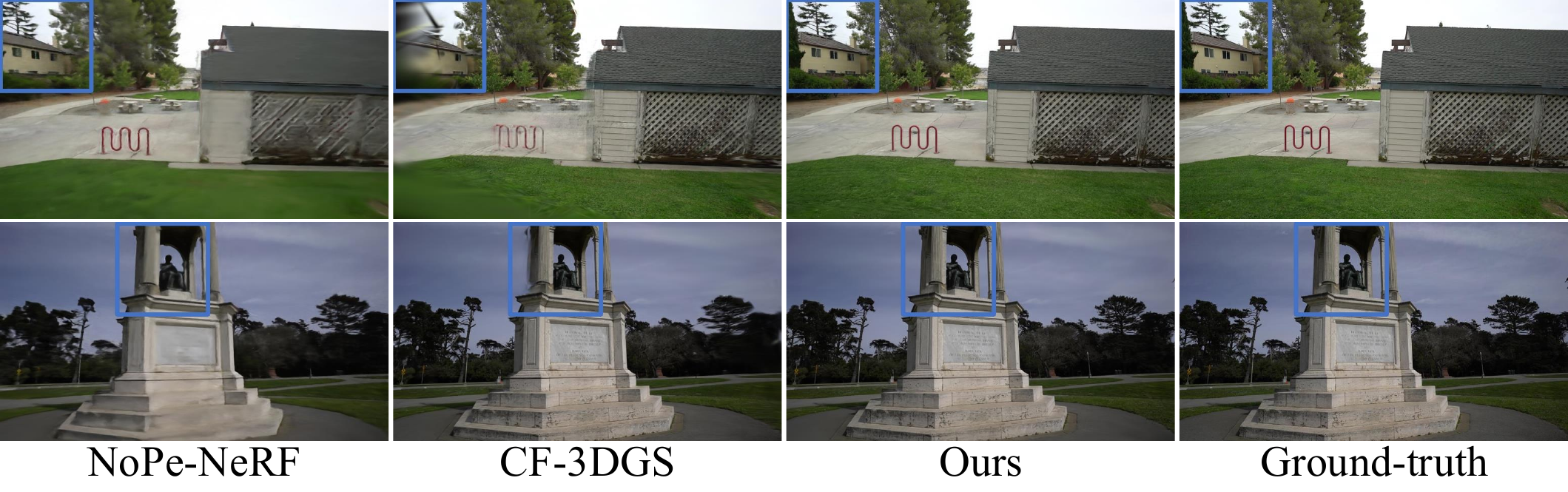}
}
\vspace{-8mm}
\caption{\textbf{Qualitative comparison on the Tanks and Temples dataset~\cite{knapitsch2017tanks}.} NoPe-NeRF~\cite{bian2023nope} produces visibly blurred results with inaccurate geometries, while CF-3DGS~\cite{fu2024colmap}, despite better sharpness, fails to reconstruct fine details accurately. In contrast, our LongSplat method achieves superior rendering quality, closely matching the ground truth with sharper textures, more accurate geometry, and consistent lighting.}
\label{fig:tnt_visual}
\end{figure}

\begin{figure*}[t]
\centering
\vspace{-6mm}
\resizebox{1.0\textwidth}{!} 
{
\includegraphics[width=\textwidth]{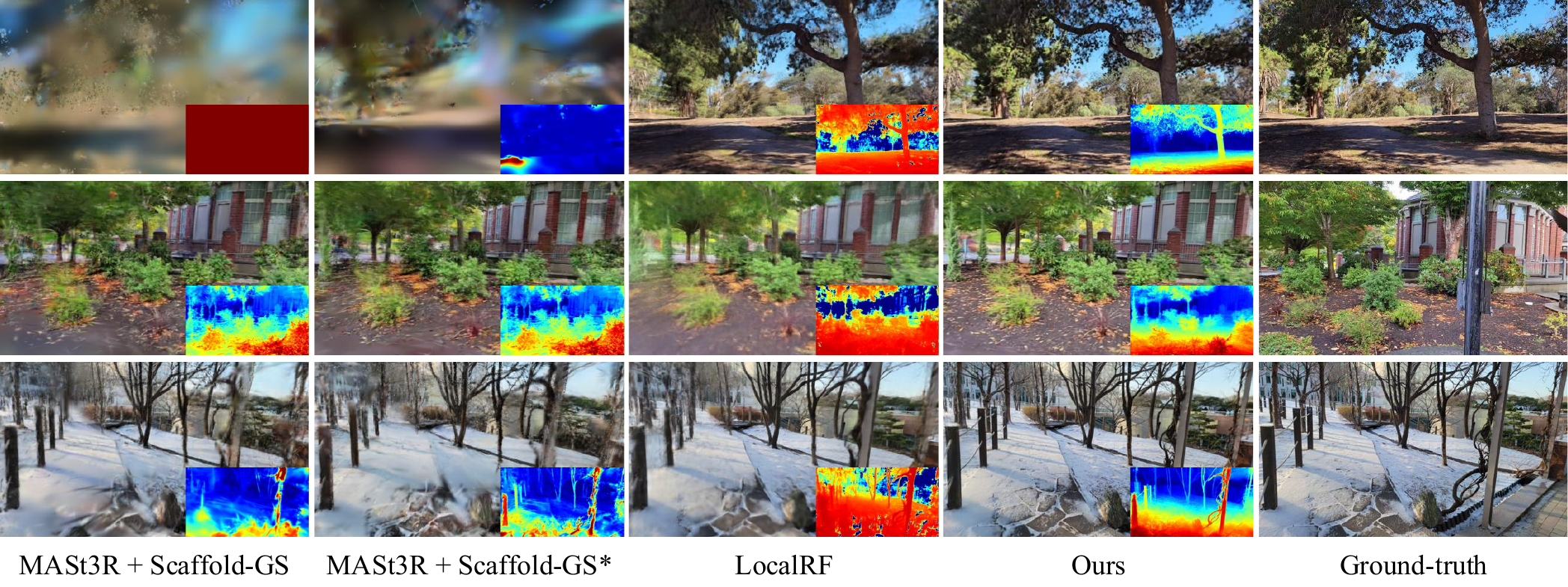}
}
\vspace{-7mm}
\caption{\textbf{Qualitative results on the Hike dataset~\cite{meuleman2023progressively}.} Compared to existing methods such as LocalRF~\cite{meuleman2023progressively} and MASt3R~\cite{leroy2024grounding}+Scaffold-GS~\cite{lu2024scaffold}, our approach significantly improves visual clarity and reconstruction fidelity, accurately capturing complex details and textures in challenging scenes captured during long, casual outdoor trajectories. Notably, our method better preserves structural details and reduces artifacts, demonstrating enhanced robustness and visual quality. ``*'': Initialized with MASt3R poses, then jointly optimized.}
\label{fig:hike_visual}
\end{figure*}

\vspace{2pt}
\noindent {\bf Implementation Details.}
We implement LongSplat based on Scaffold-GS~\cite{lu2024scaffold}, using its learning rate schedule and growing/pruning rules. Each anchor emits $k$ Gaussians predicted by a lightweight 2-layer MLP. The initial sparse voxel grid size is 0.1. Camera poses are optimized via a differentiable CUDA-accelerated rasterizer, parameterized with quaternions and translation vectors. We use 400 local, 900 global, and 10,000 refinement iterations, starting with three initial frames. The octree density thresholds for splitting and removal start at 10 and 5, progressively increasing with depth. Visibility IoU threshold is set to 0.2. All experiments are conducted on a single NVIDIA RTX 4090.


\subsection{Comparisons}

\noindent {\bf Tanks and Temples.}
We evaluate LongSplat on the Tanks and Temples dataset~\cite{knapitsch2017tanks}, a standard benchmark for novel view synthesis. As shown in~\cref{tab:tnt_average_quantitative}, LongSplat achieves state-of-the-art rendering quality (avg. PSNR: 32.83 dB) and superior camera pose estimation accuracy (lowest ATE and RPE). Qualitative results in~\cref{fig:tnt_visual} confirm sharper textures, accurate geometry, and better visual consistency compared to baselines. Please refer to the supplementary material for the full quantitative evaluation table for each scene.


\vspace{2pt}
\noindent {\bf Free Dataset.}
We evaluate LongSplat on the challenging Free dataset, achieving superior reconstruction quality as shown in~\cref{tab:free_quantitative} and~\cref{fig:free_visual}. Competing methods like CF-3DGS often face OOM issues, while LocalRF produces fragmented geometry and pose drift. Although MASt3R + Scaffold-GS avoids OOM errors, its inaccurate global pose estimates from MASt3R result in blurred renderings and structural distortions. Our method also achieves consistently lower pose errors than baselines, as shown quantitatively in~\cref{tab:free_pose_quantitative} and visually in~\cref{fig:free_traj_visual}.

\vspace{2pt}
\noindent {\bf Hike Dataset.}
We evaluate LongSplat on the challenging Hike dataset, achieving state-of-the-art reconstruction quality as shown in \cref{tab:hike_quantitative_avg} and \cref{fig:hike_visual}. Competing methods like CF-3DGS often fail due to OOM issue, LocalRF produces lower-quality reconstructions, and MASt3R struggles with long outdoor trajectories, resulting in poor reconstruction quality.

\subsection{Ablation Studies}
\begin{table}[t]
\centering
\small
\renewcommand{\arraystretch}{0.8}
\caption{
\textbf{Quantitative evaluation on the Hike dataset~\cite{meuleman2023progressively}.} Our method consistently outperforms baselines across diverse scenes with complex trajectories and extended sequences, highlighting LongSplat’s robustness and superior scene representation capability. CF-3DGS~\cite{fu2024colmap} encounters OOM in all scenes and is thus omitted.
}
\vspace{-3mm}
\label{tab:hike_quantitative_avg}
\footnotesize
\resizebox{\columnwidth}{!}{%
\begin{tabular}{l|ccc}
\toprule
Method & PSNR$\uparrow$ & SSIM$\uparrow$ & LPIPS$\downarrow$ \\
\midrule
MASt3R~\cite{leroy2024grounding} + Scaffold-GS~\cite{lu2024scaffold}   & 17.30 & 0.42 & 0.52 \\
MASt3R~\cite{leroy2024grounding} + Scaffold-GS~\cite{lu2024scaffold}*  & 17.90 & 0.44 & 0.50 \\
LocalRF~\cite{meuleman2023progressively}               & 23.56 & 0.68 & 0.29 \\
Ours                           & \textbf{25.39} & \textbf{0.81} & \textbf{0.19} \\
\bottomrule
\end{tabular}%
}
\end{table}

\begin{table}[t]
    \centering
    \small
\vspace{-2mm}
    \renewcommand{\arraystretch}{0.8}
    \caption{
    \textbf{Ablation on training components.} Removing pose estimation, global optimization, or local optimization significantly degrades performance, highlighting each module’s importance. Our full method achieves the best rendering quality and pose accuracy.
    }
    \vspace{-3mm}
    \label{tab:abalation_training_strategy}
    \resizebox{\columnwidth}{!}{
    \begin{tabular}{l|cccccc}
    \toprule
    Method & PSNR$\uparrow$ & SSIM$\uparrow$ & LPIPS$\downarrow$ & RPE$_t$$\downarrow$ & RPE$_r$$\downarrow$ & ATE$\downarrow$ \\
    \midrule
    w/o Pose estimation    & 20.19 & 0.56 & 0.51 & 0.42 & 2.71 & 0.71 \\
    w/o Global optimization& 20.50 & 0.58 & 0.41 & 0.12 & 0.50 & 0.01 \\
    w/o Local optimization & 25.94 & 0.77 & 0.28 & 0.06 & 0.31 & 0.01 \\
    w/o Refinement         & 26.08 & 0.80 & 0.25 & 0.04 & 0.22 & 0.01 \\
    Ours                   & \textbf{27.88} & \textbf{0.85} & \textbf{0.17} & \textbf{0.03} & \textbf{0.11} & \textbf{0.00} \\
    \bottomrule
    \end{tabular}%
}

\end{table}

\noindent {\bf Training Components.}
To analyze the contribution of each training component, we individually disable them and evaluate performance. As shown in~\cref{tab:abalation_training_strategy}, removing pose estimation severely harms reconstruction quality and increases pose errors (ATE: 0.71). Omitting global or local optimization also reduces performance. Our full method achieves the highest quality and minimal pose errors.

\begin{table}[t]
    \centering
    \small
    \setlength{\tabcolsep}{4pt} 
    \renewcommand{\arraystretch}{0.8}
    \caption{
    \textbf{Ablation on local window sizes.} Fixed small windows (e.g., 1-frame or 5-frame) or global optimization degrades reconstruction quality and pose accuracy. Our visibility-adaptive window dynamically selects optimal context, achieving the best balance of local detail and global consistency.
    }
    \vspace{-3mm}
    \label{tab:ablation_window_size}
    \resizebox{\columnwidth}{!}{
    \begin{tabular}{l|cccccc}
    \toprule
    Window size & PSNR $\uparrow$ & SSIM $\uparrow$ & LPIPS $\downarrow$ & RPE$_t$$\downarrow$ & RPE$_r$$\downarrow$ & ATE$\downarrow$ \\
    \midrule
    1-frame (Minimum Window)     & 26.58 & 0.80 & 0.23 & 0.05 & 0.21 & 0.01 \\
    5-frame (Fixed Window)       & 26.90 & 0.82 & 0.22 & 0.04 & 0.18 & 0.01 \\
    All Frames (Global Optimize) & 26.15 & 0.78 & 0.26 & 0.06 & 0.28 & 0.08 \\
    Ours (Visibility-Adaptive)   & \textbf{27.88} & \textbf{0.85} & \textbf{0.17} & \textbf{0.03} & \textbf{0.11} & \textbf{0.00} \\
    \bottomrule
    \end{tabular}%
    }
\end{table}

\vspace{2pt}
\noindent {\bf Local Window Sizes.}
We analyze the effect of local window size on reconstruction and pose accuracy in ~\cref{tab:ablation_window_size}. Small fixed-size windows (e.g., 1 frame) lack sufficient constraints, causing fragmentation and higher errors. Our visibility-adapted window achieves the best balance, yielding the highest reconstruction quality and lowest pose drift.

\begin{table}[t]
    \centering
    \small
    \renewcommand{\arraystretch}{0.8}
    \caption{
    \textbf{Ablation on anchor unprojection strategies.} Our Adaptive Octree method achieves the best rendering quality and lowest perceptual errors, significantly reducing memory usage (7.92$\times$ compression) compared to baselines.
}
\vspace{-3mm}
    \label{tab:ablation_anchor_voxel}
    \resizebox{\columnwidth}{!}{
    \begin{tabular}{l|ccccc}
    \toprule
    Method & PSNR$\uparrow$ & SSIM$\uparrow$ & LPIPS$\downarrow$ & Size (MB)$\downarrow$ & Compress$\uparrow$  \\
    \midrule
    Per-pixel Unprojection (Dense)   & 22.47 & 0.69 & 0.35 & 799  & 1.00x \\
    Fixed-size Voxel Unprojection    & \cellcolor{orange!25}{26.99} & \cellcolor{orange!25}{0.81} & \cellcolor{orange!25}{0.18} & 591  & 1.35x \\
    Naive Densification              & 25.73 & 0.75 & 0.31 & \cellcolor{red!25}{63}   & \cellcolor{red!25}{12.66x} \\
    Ours (Adaptive Octree)           & \cellcolor{red!25}{27.88} & \cellcolor{red!25}{0.85} & \cellcolor{red!25}{0.17} & \cellcolor{orange!25}{101} & \cellcolor{orange!25}{7.92x} \\
    \bottomrule
    \end{tabular}%
    }
\end{table}

\vspace{2pt}
\noindent {\bf Anchor Unprojection Strategies.}
We compare our adaptive octree anchor formation to (1) per-pixel initialization, (2) fixed-resolution voxels, and (3) naïve densification in \cref{tab:ablation_anchor_voxel}. Our method achieves superior reconstruction quality with significantly reduced memory usage (7.92$\times$ compression).

\begin{table}[t]
    \centering
    \small
    \footnotesize
    \caption{\textbf{Comparison of training efficiency on the Free dataset.} Our method significantly reduces training time and achieves dramatically higher throughput (FPS) while simultaneously maintaining a compact model size compared to state-of-the-art approaches.}
    \vspace{-3mm}
    \label{tab:training_compare}
    \resizebox{0.8\columnwidth}{!}{
    \begin{tabular}{l|ccc}
    \toprule
    Method & FPS $\uparrow$ & Training time $\downarrow$ & Size (MB) $\downarrow$ \\
    \midrule
    NoPe-NeRF~\cite{bian2023nope}   & 0.29 & 36 hr & \cellcolor{red!25}7 \\
    LocalRF~\cite{meuleman2023progressively} & 1.17 & 14 hr & 1080 \\
    CF-3DGS~\cite{fu2024colmap} & \cellcolor{orange!25}9.81 & \cellcolor{orange!25}2 hr & 1966 \\
    Ours    & \cellcolor{red!25}281.71 & \cellcolor{red!25}1 hr & \cellcolor{orange!25}101 \\
    \bottomrule
    \end{tabular}%
    }
\end{table}

\vspace{2pt}
\noindent {\bf Training Efficiency.}
We evaluate the computational efficiency of LongSplat (\cref{tab:training_compare}), which achieves 281.71 FPS and trains in just 1 hour on an NVIDIA RTX 4090, nearly 30× faster than LocalRF. Our method also significantly reduces the model size to approximately 101 MB.

\begin{figure}[t]
\centering
\footnotesize
\renewcommand{\arraystretch}{0.1}
\resizebox{\columnwidth}{!}{
\begin{tabular}{ccc}
\includegraphics[width=0.33\columnwidth]{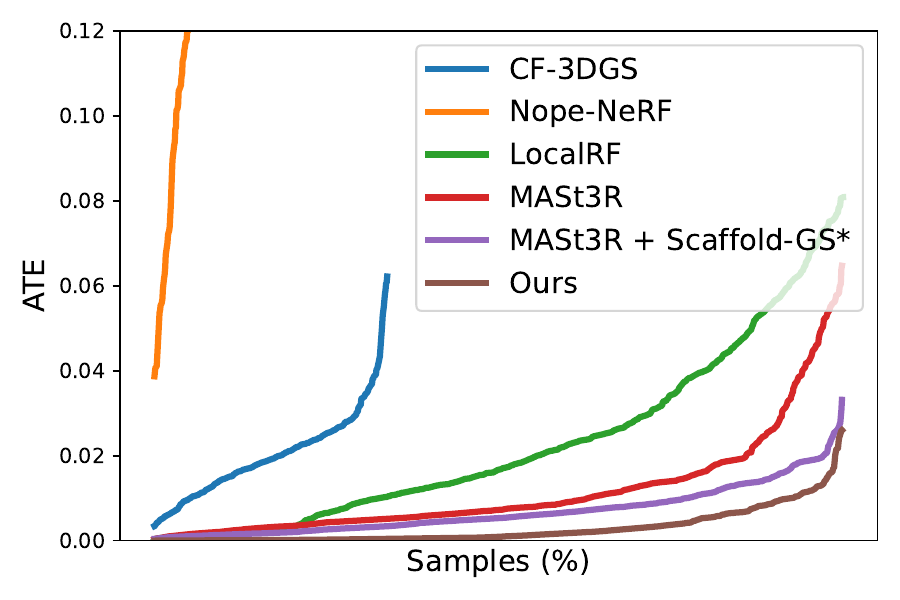} & 
\includegraphics[width=0.33\columnwidth]{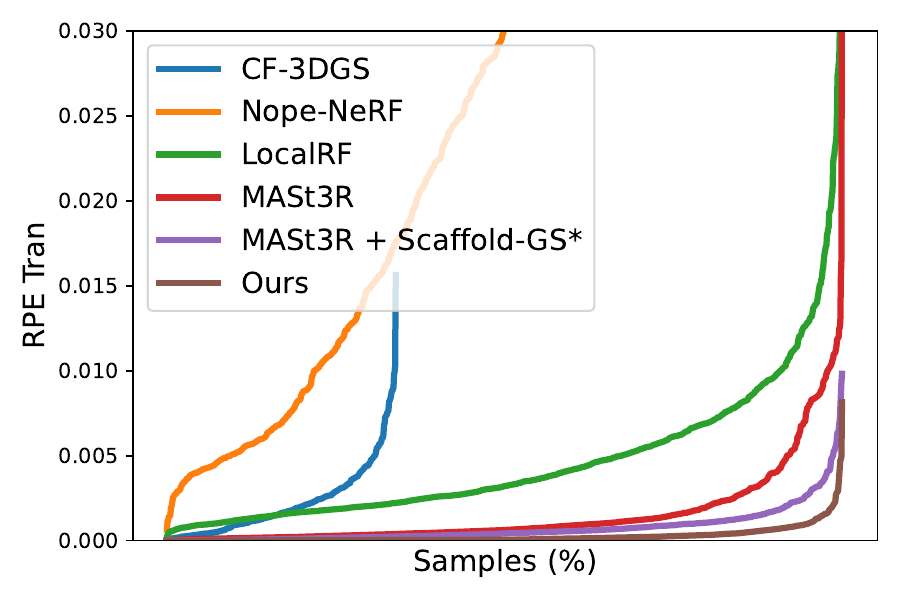} &
\includegraphics[width=0.33\columnwidth]{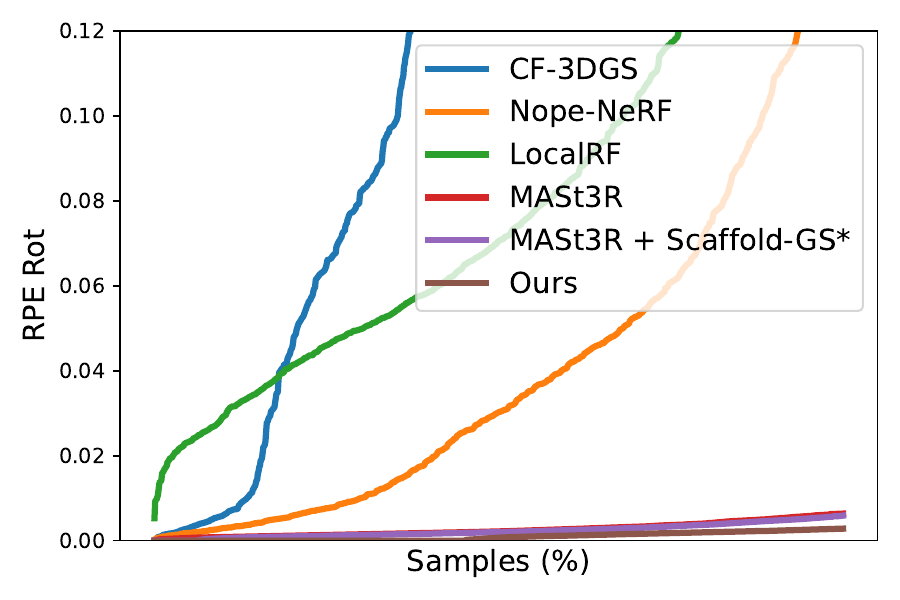} \\
 {(a) ATE} &  {(b) $\text{RPE}_t$} & {(c) $\text{RPE}_r$}
\end{tabular}
}
\vspace{-3mm}
\caption{
\textbf{Robustness analysis on camera pose estimation (Free dataset~\cite{wang2023f2}).} We plot cumulative error distributions for ATE, RPE translation, and rotation. Our method consistently achieves lower errors compared to existing methods, demonstrating superior robustness and reduced pose drift.
}
\label{fig:pose_robustness}
\end{figure}

\vspace{2pt}
\noindent {\bf Robustness Analysis of Camera Pose Estimation.}
We further analyze robustness by comparing cumulative error distributions for ATE and RPE (translation and rotation) in~\cref{fig:pose_robustness}. LongSplat achieves consistently lower errors than baselines, effectively minimizing drift and maintaining stable trajectories, highlighting the advantage of our incremental optimization and robust tracking.



\section{Conclusion}
\label{sec:conclusion}
We present LongSplat, a robust unposed 3D Gaussian Splatting framework for casual long videos. It integrates incremental joint optimization, a robust tracking module, and adaptive octree anchors, significantly improving pose accuracy, reconstruction quality, and memory efficiency. Extensive experiments confirm that LongSplat consistently outperforms state-of-the-art approaches. Future work includes handling dynamic scenes and enhancing pose estimation robustness.

\vspace{2pt}
\noindent {\bf Limitations.} LongSplat shares common limitations of unposed reconstruction methods, assuming static scenes and fixed intrinsics, making it unsuitable for dynamic objects or varying focal lengths.


\paragraph{Acknowledgements.}
This work was supported by NVIDIA Taiwan AI Research \& Development Center (TRDC). This research was funded by the National Science and Technology Council, Taiwan, under Grants NSTC 112-2222-E-A49-004-MY2 and 113-2628-E-A49-023-. Yu-Lun Liu acknowledges the Yushan Young Fellow Program by the MOE in Taiwan.

{\small
\bibliographystyle{ieeenat_fullname}
\bibliography{11_references}

\begin{thebibliography}{77}
\providecommand{\natexlab}[1]{#1}
\providecommand{\url}[1]{\texttt{#1}}
\expandafter\ifx\csname urlstyle\endcsname\relax
  \providecommand{\doi}[1]{doi: #1}\else
  \providecommand{\doi}{doi: \begingroup \urlstyle{rm}\Url}\fi

\bibitem[Attal et~al.(2023)Attal, Huang, Richardt, Zollh\"ofer, Kopf, O{\textquoteright}Toole, and Kim]{Attal_2023_CVPR}
Benjamin Attal, Jia-Bin Huang, Christian Richardt, Michael Zollh\"ofer, Johannes Kopf, Matthew O{\textquoteright}Toole, and Changil Kim.
\newblock Hyperreel: High-fidelity 6-dof video with ray-conditioned sampling.
\newblock In \emph{CVPR}, 2023.

\bibitem[Barron et~al.(2021)Barron, Mildenhall, Tancik, Hedman, Martin-Brualla, and Srinivasan]{barron2021mip}
Jonathan~T Barron, Ben Mildenhall, Matthew Tancik, Peter Hedman, Ricardo Martin-Brualla, and Pratul~P Srinivasan.
\newblock Mip-nerf: A multiscale representation for anti-aliasing neural radiance fields.
\newblock In \emph{ICCV}, 2021.

\bibitem[Barron et~al.(2022)Barron, Mildenhall, Verbin, Srinivasan, and Hedman]{barron2022mip}
Jonathan~T Barron, Ben Mildenhall, Dor Verbin, Pratul~P Srinivasan, and Peter Hedman.
\newblock Mip-nerf 360: Unbounded anti-aliased neural radiance fields.
\newblock In \emph{CVPR}, 2022.

\bibitem[Barron et~al.(2023)Barron, Mildenhall, Verbin, Srinivasan, and Hedman]{barron2023zip}
Jonathan~T Barron, Ben Mildenhall, Dor Verbin, Pratul~P Srinivasan, and Peter Hedman.
\newblock Zip-nerf: Anti-aliased grid-based neural radiance fields.
\newblock \emph{arXiv preprint arXiv:2304.06706}, 2023.

\bibitem[Bian et~al.(2023)Bian, Wang, Li, Bian, and Prisacariu]{bian2023nope}
Wenjing Bian, Zirui Wang, Kejie Li, Jia-Wang Bian, and Victor~Adrian Prisacariu.
\newblock Nope-nerf: Optimising neural radiance field with no pose prior.
\newblock In \emph{CVPR}, 2023.

\bibitem[Buehler et~al.(2001)Buehler, Bosse, McMillan, Gortler, and Cohen]{buehler2001unstructured}
Chris Buehler, Michael Bosse, Leonard McMillan, Steven Gortler, and Michael Cohen.
\newblock Unstructured lumigraph rendering.
\newblock In \emph{SIGGRAPH}, 2001.

\bibitem[Chen et~al.(2024)Chen, Chiu, and Liu]{chen2024improving}
Bo-Yu Chen, Wei-Chen Chiu, and Yu-Lun Liu.
\newblock Improving robustness for joint optimization of camera pose and decomposed low-rank tensorial radiance fields.
\newblock In \emph{AAAI}, 2024.

\bibitem[Chen and Williams(1993)]{chen1993view}
Shenchang~Eric Chen and Lance Williams.
\newblock View interpolation for image synthesis.
\newblock In \emph{SIGGRAPH}, 1993.

\bibitem[Cheng et~al.(2023)Cheng, Esteves, Jampani, Kar, Maji, and Makadia]{cheng2023lu}
Zezhou Cheng, Carlos Esteves, Varun Jampani, Abhishek Kar, Subhransu Maji, and Ameesh Makadia.
\newblock Lu-nerf: Scene and pose estimation by synchronizing local unposed nerfs.
\newblock \emph{arXiv preprint arXiv:2306.05410}, 2023.

\bibitem[Chng et~al.(2022)Chng, Ramasinghe, Sherrah, and Lucey]{chng2022garf}
Shin-Fang Chng, Sameera Ramasinghe, Jamie Sherrah, and Simon Lucey.
\newblock Garf: Gaussian activated radiance fields for high fidelity reconstruction and pose estimation.
\newblock \emph{arXiv e-prints}, 2022.

\bibitem[Cong et~al.(2025)Cong, Wang, Lei, Stearns, Cai, Wang, Ranjan, Feiszli, Guibas, Wang, Wang, and Fan]{cong2025videolifter}
Wenyan Cong, Kevin Wang, Jiahui Lei, Colton Stearns, Yuanhao Cai, Dilin Wang, Rakesh Ranjan, Matt Feiszli, Leonidas Guibas, Zhangyang Wang, Weiyao Wang, and Zhiwen Fan.
\newblock Videolifter: Lifting videos to 3d with fast hierarchical stereo alignment, 2025.

\bibitem[Debevec et~al.(1996)Debevec, Taylor, and Malik]{debevec1996modeling}
Paul~E Debevec, Camillo~J Taylor, and Jitendra Malik.
\newblock Modeling and rendering architecture from photographs: A hybrid geometry-and image-based approach.
\newblock In \emph{SIGGRAPH}, 1996.

\bibitem[Fan et~al.(2025)Fan, Chang, Liu, Lee, Huang, Tseng, and Liu]{fan2025spectromotion}
Cheng-De Fan, Chen-Wei Chang, Yi-Ruei Liu, Jie-Ying Lee, Jiun-Long Huang, Yu-Chee Tseng, and Yu-Lun Liu.
\newblock Spectromotion: Dynamic 3d reconstruction of specular scenes.
\newblock In \emph{CVPR}, 2025.

\bibitem[Fu et~al.(2024)Fu, Liu, Kulkarni, Kautz, Efros, and Wang]{fu2024colmap}
Yang Fu, Sifei Liu, Amey Kulkarni, Jan Kautz, Alexei~A Efros, and Xiaolong Wang.
\newblock Colmap-free 3d gaussian splatting.
\newblock In \emph{CVPR}, 2024.

\bibitem[Garbin et~al.(2021)Garbin, Kowalski, Johnson, Shotton, and Valentin]{garbin2021fastnerf}
Stephan~J Garbin, Marek Kowalski, Matthew Johnson, Jamie Shotton, and Julien Valentin.
\newblock Fastnerf: High-fidelity neural rendering at 200fps.
\newblock In \emph{ICCV}, 2021.

\bibitem[Hartley and Zisserman(2003)]{hartley2003multiple}
Richard Hartley and Andrew Zisserman.
\newblock \emph{Multiple view geometry in computer vision}.
\newblock 2003.

\bibitem[Hoiem et~al.(2005)Hoiem, Efros, and Hebert]{hoiem2005automatic}
Derek Hoiem, Alexei~A Efros, and Martial Hebert.
\newblock Automatic photo pop-up.
\newblock In \emph{ACM SIGGRAPH 2005 Papers}, 2005.

\bibitem[Horry et~al.(1997)Horry, Anjyo, and Arai]{horry1997tour}
Youichi Horry, Ken-Ichi Anjyo, and Kiyoshi Arai.
\newblock Tour into the picture: using a spidery mesh interface to make animation from a single image.
\newblock In \emph{Proceedings of the 24th annual conference on Computer graphics and interactive techniques}, 1997.

\bibitem[Hou et~al.(2025)Hou, Hsu, Huang, Shen, Sun, Sun, Chang, Liu, and Lee]{hou20253d}
Hao-Yu Hou, Chia-Chi Hsu, Yu-Chen Huang, Mu-Yi Shen, Wei-Fang Sun, Cheng Sun, Chia-Che Chang, Yu-Lun Liu, and Chun-Yi Lee.
\newblock 3d gaussian splatting with grouped uncertainty for unconstrained images.
\newblock In \emph{ICASSP}, 2025.

\bibitem[Hu et~al.(2020)Hu, Ravi, Berg, and Pathak]{hu2020worldsheet}
Ronghang Hu, Nikhila Ravi, Alex Berg, and Deepak Pathak.
\newblock Worldsheet: Wrapping the world in a 3d sheet for view synthesis from a single image.
\newblock In \emph{ICCV}, 2020.

\bibitem[Ji and Yao(2024)]{ji2024sfm}
Bo Ji and Angela Yao.
\newblock Sfm-free 3d gaussian splatting via hierarchical training.
\newblock \emph{arXiv preprint arXiv:2412.01553}, 2024.

\bibitem[Kerbl et~al.(2023)Kerbl, Kopanas, Leimk{\"u}hler, and Drettakis]{kerbl20233d}
Bernhard Kerbl, Georgios Kopanas, Thomas Leimk{\"u}hler, and George Drettakis.
\newblock 3d gaussian splatting for real-time radiance field rendering.
\newblock \emph{ACM TOG}, 2023.

\bibitem[Kerbl et~al.(2024)Kerbl, Meuleman, Kopanas, Wimmer, Lanvin, and Drettakis]{kerbl2024hierarchical}
Bernhard Kerbl, Andreas Meuleman, Georgios Kopanas, Michael Wimmer, Alexandre Lanvin, and George Drettakis.
\newblock A hierarchical 3d gaussian representation for real-time rendering of very large datasets.
\newblock \emph{ACM TOG}, 2024.

\bibitem[Kim et~al.(2022)Kim, Seo, and Han]{kim2022infonerf}
Mijeong Kim, Seonguk Seo, and Bohyung Han.
\newblock Infonerf: Ray entropy minimization for few-shot neural volume rendering.
\newblock In \emph{CVPR}, 2022.

\bibitem[Knapitsch et~al.(2017)Knapitsch, Park, Zhou, and Koltun]{knapitsch2017tanks}
Arno Knapitsch, Jaesik Park, Qian-Yi Zhou, and Vladlen Koltun.
\newblock Tanks and temples: Benchmarking large-scale scene reconstruction.
\newblock \emph{ACM TOG}, 2017.

\bibitem[Kopf et~al.(2014)Kopf, Cohen, and Szeliski]{Kopf2014}
Johannes Kopf, Michael~F. Cohen, and Richard Szeliski.
\newblock First-person hyper-lapse videos.
\newblock \emph{ACM TOG}, 2014.

\bibitem[Leroy et~al.(2024)Leroy, Cabon, and Revaud]{leroy2024grounding}
Vincent Leroy, Yohann Cabon, and J{\'e}r{\^o}me Revaud.
\newblock Grounding image matching in 3d with mast3r.
\newblock In \emph{ECCV}, 2024.

\bibitem[Li et~al.(2013)Li, Hang, and Liu]{li2013virtual}
Du-Hsiu Li, Hsueh-Ming Hang, and Yu-Lun Liu.
\newblock Virtual view synthesis using backward depth warping algorithm.
\newblock In \emph{PCS}, 2013.

\bibitem[Li et~al.(2021)Li, Feng, She, Ding, Wang, and Lee]{li2021mine}
Jiaxin Li, Zijian Feng, Qi She, Henghui Ding, Changhu Wang, and Gim~Hee Lee.
\newblock Mine: Towards continuous depth mpi with nerf for novel view synthesis.
\newblock In \emph{ICCV}, 2021.

\bibitem[Li et~al.(2024)Li, Ku, Yen, Liu, Liu, Chen, Kuo, and Sun]{li2024genrc}
Ming-Feng Li, Yueh-Feng Ku, Hong-Xuan Yen, Chi Liu, Yu-Lun Liu, Albert~YC Chen, Cheng-Hao Kuo, and Min Sun.
\newblock Genrc: Generative 3d room completion from sparse image collections.
\newblock In \emph{ECCV}, 2024.

\bibitem[Lin et~al.(2021)Lin, Ma, Torralba, and Lucey]{lin2021barf}
Chen-Hsuan Lin, Wei-Chiu Ma, Antonio Torralba, and Simon Lucey.
\newblock Barf: Bundle-adjusting neural radiance fields.
\newblock In \emph{ICCV}, 2021.

\bibitem[Lin et~al.(2025)Lin, Wu, Yeh, Yen, Sun, and Liu]{lin2024frugalnerf}
Chin-Yang Lin, Chung-Ho Wu, Chang-Han Yeh, Shih-Han Yen, Cheng Sun, and Yu-Lun Liu.
\newblock Frugalnerf: Fast convergence for few-shot novel view synthesis without learned priors.
\newblock \emph{CVPR}, 2025.

\bibitem[Lin et~al.(2024)Lin, Li, Tang, Liu, Liu, Liu, Lu, Wu, Xu, Yan, et~al.]{lin2024vastgaussian}
Jiaqi Lin, Zhihao Li, Xiao Tang, Jianzhuang Liu, Shiyong Liu, Jiayue Liu, Yangdi Lu, Xiaofei Wu, Songcen Xu, Youliang Yan, et~al.
\newblock Vastgaussian: Vast 3d gaussians for large scene reconstruction.
\newblock In \emph{CVPR}, 2024.

\bibitem[Liu et~al.(2020)Liu, Gu, Zaw~Lin, Chua, and Theobalt]{liu2020neural}
Lingjie Liu, Jiatao Gu, Kyaw Zaw~Lin, Tat-Seng Chua, and Christian Theobalt.
\newblock Neural sparse voxel fields.
\newblock In \emph{NeurIPS}, 2020.

\bibitem[Liu et~al.(2023)Liu, Gao, Meuleman, Tseng, Saraf, Kim, Chuang, Kopf, and Huang]{liu2023robust}
Yu-Lun Liu, Chen Gao, Andreas Meuleman, Hung-Yu Tseng, Ayush Saraf, Changil Kim, Yung-Yu Chuang, Johannes Kopf, and Jia-Bin Huang.
\newblock Robust dynamic radiance fields.
\newblock In \emph{CVPR}, 2023.

\bibitem[Lu et~al.(2024)Lu, Yu, Xu, Xiangli, Wang, Lin, and Dai]{lu2024scaffold}
Tao Lu, Mulin Yu, Linning Xu, Yuanbo Xiangli, Limin Wang, Dahua Lin, and Bo Dai.
\newblock Scaffold-gs: Structured 3d gaussians for view-adaptive rendering.
\newblock In \emph{CVPR}, 2024.

\bibitem[Luiten et~al.(2023)Luiten, Kopanas, Leibe, and Ramanan]{luiten2023dynamic}
Jonathon Luiten, Georgios Kopanas, Bastian Leibe, and Deva Ramanan.
\newblock Dynamic 3d gaussians: Tracking by persistent dynamic view synthesis.
\newblock \emph{arXiv preprint arXiv:2308.09713}, 2023.

\bibitem[Ma et~al.(2024)Ma, Liu, Wang, Liu, Liu, and Wang]{ma2024humannerf}
Caoyuan Ma, Yu-Lun Liu, Zhixiang Wang, Wu Liu, Xinchen Liu, and Zheng Wang.
\newblock Humannerf-se: A simple yet effective approach to animate humannerf with diverse poses.
\newblock In \emph{CVPR}, 2024.

\bibitem[Meuleman et~al.(2023)Meuleman, Liu, Gao, Huang, Kim, Kim, and Kopf]{meuleman2023progressively}
Andreas Meuleman, Yu-Lun Liu, Chen Gao, Jia-Bin Huang, Changil Kim, Min~H Kim, and Johannes Kopf.
\newblock Progressively optimized local radiance fields for robust view synthesis.
\newblock In \emph{CVPR}, 2023.

\bibitem[Mildenhall et~al.(2021)Mildenhall, Srinivasan, Tancik, Barron, Ramamoorthi, and Ng]{mildenhall2021nerf}
Ben Mildenhall, Pratul~P Srinivasan, Matthew Tancik, Jonathan~T Barron, Ravi Ramamoorthi, and Ren Ng.
\newblock Nerf: Representing scenes as neural radiance fields for view synthesis.
\newblock \emph{Communications of the ACM}, 2021.

\bibitem[M\"uller et~al.(2022)M\"uller, Evans, Schied, and Keller]{mueller2022instant}
Thomas M\"uller, Alex Evans, Christoph Schied, and Alexander Keller.
\newblock Instant neural graphics primitives with a multiresolution hash encoding.
\newblock \emph{ACM TOG}, 2022.

\bibitem[Mur-Artal et~al.(2015)Mur-Artal, Montiel, and Tardos]{mur2015orb}
Raul Mur-Artal, Jose Maria~Martinez Montiel, and Juan~D Tardos.
\newblock Orb-slam: a versatile and accurate monocular slam system.
\newblock \emph{IEEE transactions on robotics}, 2015.

\bibitem[Niemeyer et~al.(2022)Niemeyer, Barron, Mildenhall, Sajjadi, Geiger, and Radwan]{niemeyer2022regnerf}
Michael Niemeyer, Jonathan~T Barron, Ben Mildenhall, Mehdi~SM Sajjadi, Andreas Geiger, and Noha Radwan.
\newblock Regnerf: Regularizing neural radiance fields for view synthesis from sparse inputs.
\newblock In \emph{CVPR}, 2022.

\bibitem[Ran et~al.(2024)Ran, Li, Ye, Huo, Bai, Sun, and Chen]{ran2024ct}
Yunlong Ran, Yanxu Li, Qi Ye, Yuchi Huo, Zechun Bai, Jiahao Sun, and Jiming Chen.
\newblock Ct-nerf: Incremental optimizing neural radiance field and poses with complex trajectory.
\newblock \emph{arXiv preprint arXiv:2404.13896}, 2024.

\bibitem[Reiser et~al.(2021)Reiser, Peng, Liao, and Geiger]{reiser2021kilonerf}
Christian Reiser, Songyou Peng, Yiyi Liao, and Andreas Geiger.
\newblock Kilonerf: Speeding up neural radiance fields with thousands of tiny mlps.
\newblock In \emph{ICCV}, 2021.

\bibitem[Reizenstein et~al.(2021)Reizenstein, Shapovalov, Henzler, Sbordone, Labatut, and Novotny]{reizenstein2021common}
Jeremy Reizenstein, Roman Shapovalov, Philipp Henzler, Luca Sbordone, Patrick Labatut, and David Novotny.
\newblock Common objects in 3d: Large-scale learning and evaluation of real-life 3d category reconstruction.
\newblock In \emph{ICCV}, 2021.

\bibitem[Ren et~al.(2024)Ren, Jiang, Lu, Yu, Xu, Ni, and Dai]{ren2024octree}
Kerui Ren, Lihan Jiang, Tao Lu, Mulin Yu, Linning Xu, Zhangkai Ni, and Bo Dai.
\newblock Octree-gs: Towards consistent real-time rendering with lod-structured 3d gaussians.
\newblock \emph{arXiv preprint arXiv:2403.17898}, 2024.

\bibitem[Riegler and Koltun(2020)]{Riegler2020FVS}
Gernot Riegler and Vladlen Koltun.
\newblock Free view synthesis.
\newblock In \emph{ECCV}, 2020.

\bibitem[Riegler and Koltun(2021)]{riegler2021stable}
Gernot Riegler and Vladlen Koltun.
\newblock Stable view synthesis.
\newblock In \emph{CVPR}, 2021.

\bibitem[{Sara Fridovich-Keil and Alex Yu} et~al.(2022){Sara Fridovich-Keil and Alex Yu}, Tancik, Chen, Recht, and Kanazawa]{yu_and_fridovichkeil2021plenoxels}
{Sara Fridovich-Keil and Alex Yu}, Matthew Tancik, Qinhong Chen, Benjamin Recht, and Angjoo Kanazawa.
\newblock Plenoxels: Radiance fields without neural networks.
\newblock In \emph{CVPR}, 2022.

\bibitem[Schonberger and Frahm(2016)]{schonberger2016structure}
Johannes~L Schonberger and Jan-Michael Frahm.
\newblock Structure-from-motion revisited.
\newblock In \emph{CVPR}, 2016.

\bibitem[Shen et~al.(2024)Shen, Hsu, Hou, Huang, Sun, Chang, Liu, and Lee]{shen2024driveenv}
Mu-Yi Shen, Chia-Chi Hsu, Hao-Yu Hou, Yu-Chen Huang, Wei-Fang Sun, Chia-Che Chang, Yu-Lun Liu, and Chun-Yi Lee.
\newblock Driveenv-nerf: Exploration of a nerf-based autonomous driving environment for real-world performance validation.
\newblock \emph{arXiv preprint arXiv:2403.15791}, 2024.

\bibitem[Su et~al.(2024)Su, Hu, Tsai, Lee, Lin, and Liu]{su2024boostmvsnerfs}
Chih-Hai Su, Chih-Yao Hu, Shr-Ruei Tsai, Jie-Ying Lee, Chin-Yang Lin, and Yu-Lun Liu.
\newblock Boostmvsnerfs: Boosting mvs-based nerfs to generalizable view synthesis in large-scale scenes.
\newblock In \emph{ACM SIGGRAPH 2024 Conference Papers}, 2024.

\bibitem[Sun et~al.(2022)Sun, Sun, and Chen]{sun2022direct}
Cheng Sun, Min Sun, and Hwann-Tzong Chen.
\newblock Direct voxel grid optimization: Super-fast convergence for radiance fields reconstruction.
\newblock In \emph{CVPR}, 2022.

\bibitem[Suzuki(2024)]{suzuki2024fed3dgs}
Teppei Suzuki.
\newblock Fed3dgs: Scalable 3d gaussian splatting with federated learning.
\newblock \emph{arXiv preprint arXiv:2403.11460}, 2024.

\bibitem[Taketomi et~al.(2017)Taketomi, Uchiyama, and Ikeda]{taketomi2017visual}
Takafumi Taketomi, Hideaki Uchiyama, and Sei Ikeda.
\newblock Visual slam algorithms: A survey from 2010 to 2016.
\newblock \emph{IPSJ Transactions on Computer Vision and Applications}, 2017.

\bibitem[Tancik et~al.(2022)Tancik, Casser, Yan, Pradhan, Mildenhall, Srinivasan, Barron, and Kretzschmar]{tancik2022block}
Matthew Tancik, Vincent Casser, Xinchen Yan, Sabeek Pradhan, Ben Mildenhall, Pratul~P Srinivasan, Jonathan~T Barron, and Henrik Kretzschmar.
\newblock Block-nerf: Scalable large scene neural view synthesis.
\newblock In \emph{CVPR}, 2022.

\bibitem[Tucker and Snavely(2020)]{tucker2020single}
Richard Tucker and Noah Snavely.
\newblock Single-view view synthesis with multiplane images.
\newblock In \emph{CVPR}, 2020.

\bibitem[Verbin et~al.(2022)Verbin, Hedman, Mildenhall, Zickler, Barron, and Srinivasan]{verbin2022ref}
Dor Verbin, Peter Hedman, Ben Mildenhall, Todd Zickler, Jonathan~T Barron, and Pratul~P Srinivasan.
\newblock Ref-nerf: Structured view-dependent appearance for neural radiance fields.
\newblock In \emph{CVPR}, 2022.

\bibitem[Wang and Agapito(2024)]{wang2024spann3r}
Hengyi Wang and Lourdes Agapito.
\newblock 3d reconstruction with spatial memory.
\newblock \emph{arXiv preprint arXiv:2408.16061}, 2024.

\bibitem[Wang et~al.(2023)Wang, Liu, Chen, Liu, Liu, Komura, Theobalt, and Wang]{wang2023f2}
Peng Wang, Yuan Liu, Zhaoxi Chen, Lingjie Liu, Ziwei Liu, Taku Komura, Christian Theobalt, and Wenping Wang.
\newblock F2-nerf: Fast neural radiance field training with free camera trajectories.
\newblock In \emph{CVPR}, 2023.

\bibitem[Wang et~al.(2025)Wang, Zhang, Holynski, Efros, and Kanazawa]{wang2025continuous}
Qianqian Wang, Yifei Zhang, Aleksander Holynski, Alexei~A Efros, and Angjoo Kanazawa.
\newblock Continuous 3d perception model with persistent state.
\newblock \emph{arXiv preprint arXiv:2501.12387}, 2025.

\bibitem[Wang et~al.(2024)Wang, Leroy, Cabon, Chidlovskii, and Revaud]{wang2024dust3r}
Shuzhe Wang, Vincent Leroy, Yohann Cabon, Boris Chidlovskii, and Jerome Revaud.
\newblock Dust3r: Geometric 3d vision made easy.
\newblock In \emph{CVPR}, 2024.

\bibitem[Wang et~al.(2004)Wang, Bovik, Sheikh, and Simoncelli]{wang2004image}
Zhou Wang, Alan~C Bovik, Hamid~R Sheikh, and Eero~P Simoncelli.
\newblock Image quality assessment: from error visibility to structural similarity.
\newblock \emph{IEEE TIP}, 2004.

\bibitem[Wang et~al.(2021)Wang, Wu, Xie, Chen, and Prisacariu]{wang2021nerfmm}
Zirui Wang, Shangzhe Wu, Weidi Xie, Min Chen, and Victor~Adrian Prisacariu.
\newblock Ne{RF}$--$: Neural radiance fields without known camera parameters.
\newblock \emph{arXiv preprint arXiv:2102.07064}, 2021.

\bibitem[Xia et~al.(2022)Xia, Tang, Timofte, and Van~Gool]{xia2022sinerf}
Yitong Xia, Hao Tang, Radu Timofte, and Luc Van~Gool.
\newblock Sinerf: Sinusoidal neural radiance fields for joint pose estimation and scene reconstruction.
\newblock 2022.

\bibitem[Xu et~al.(2022{\natexlab{a}})Xu, Jiang, Wang, Fan, Shi, and Wang]{xu2022sinnerf}
Dejia Xu, Yifan Jiang, Peihao Wang, Zhiwen Fan, Humphrey Shi, and Zhangyang Wang.
\newblock Sinnerf: Training neural radiance fields on complex scenes from a single image.
\newblock In \emph{ECCV}, 2022{\natexlab{a}}.

\bibitem[Xu et~al.(2022{\natexlab{b}})Xu, Xu, Philip, Bi, Shu, Sunkavalli, and Neumann]{xu2022point}
Qiangeng Xu, Zexiang Xu, Julien Philip, Sai Bi, Zhixin Shu, Kalyan Sunkavalli, and Ulrich Neumann.
\newblock Point-nerf: Point-based neural radiance fields.
\newblock In \emph{CVPR}, 2022{\natexlab{b}}.

\bibitem[Yang et~al.(2023)Yang, Pavone, and Wang]{yang2023freenerf}
Jiawei Yang, Marco Pavone, and Yue Wang.
\newblock Freenerf: Improving few-shot neural rendering with free frequency regularization.
\newblock In \emph{CVPR}, 2023.

\bibitem[Yang et~al.(2025)Yang, Sax, Liang, Henaff, Tang, Cao, Chai, Meier, and Feiszli]{yang2025fast3r}
Jianing Yang, Alexander Sax, Kevin~J Liang, Mikael Henaff, Hao Tang, Ang Cao, Joyce Chai, Franziska Meier, and Matt Feiszli.
\newblock Fast3r: Towards 3d reconstruction of 1000+ images in one forward pass.
\newblock \emph{arXiv preprint arXiv:2501.13928}, 2025.

\bibitem[Yen-Chen et~al.(2021)Yen-Chen, Florence, Barron, Rodriguez, Isola, and Lin]{yen2021inerf}
Lin Yen-Chen, Pete Florence, Jonathan~T Barron, Alberto Rodriguez, Phillip Isola, and Tsung-Yi Lin.
\newblock inerf: Inverting neural radiance fields for pose estimation.
\newblock In \emph{IROS}, 2021.

\bibitem[Yu et~al.(2021)Yu, Li, Tancik, Li, Ng, and Kanazawa]{yu2021plenoctrees}
Alex Yu, Ruilong Li, Matthew Tancik, Hao Li, Ren Ng, and Angjoo Kanazawa.
\newblock Plenoctrees for real-time rendering of neural radiance fields.
\newblock In \emph{ICCV}, 2021.

\bibitem[Zhan et~al.(2025)Zhan, Ho, Yang, Chen, Chiang, Liu, and Peng]{zhan2025cat}
Yu-Ting Zhan, Cheng-Yuan Ho, Hebi Yang, Yi-Hsin Chen, Jui~Chiu Chiang, Yu-Lun Liu, and Wen-Hsiao Peng.
\newblock Cat-3dgs: A context-adaptive triplane approach to rate-distortion-optimized 3dgs compression.
\newblock \emph{arXiv preprint arXiv:2503.00357}, 2025.

\bibitem[Zhang et~al.(2020)Zhang, Riegler, Snavely, and Koltun]{zhang2020nerf++}
Kai Zhang, Gernot Riegler, Noah Snavely, and Vladlen Koltun.
\newblock Nerf++: Analyzing and improving neural radiance fields.
\newblock \emph{arXiv:2010.07492}, 2020.

\bibitem[Zhang et~al.(2022)Zhang, Baek, Rusinkiewicz, and Heide]{zhang2022differentiable}
Qiang Zhang, Seung-Hwan Baek, Szymon Rusinkiewicz, and Felix Heide.
\newblock Differentiable point-based radiance fields for efficient view synthesis.
\newblock In \emph{SIGGRAPH Asia 2022 Conference Papers}, 2022.

\bibitem[Zhang et~al.(2018)Zhang, Isola, Efros, Shechtman, and Wang]{zhang2018unreasonable}
Richard Zhang, Phillip Isola, Alexei~A Efros, Eli Shechtman, and Oliver Wang.
\newblock The unreasonable effectiveness of deep features as a perceptual metric.
\newblock In \emph{CVPR}, 2018.

\bibitem[Zhou et~al.(2018)Zhou, Tucker, Flynn, Fyffe, and Snavely]{zhou2018stereo}
Tinghui Zhou, Richard Tucker, John Flynn, Graham Fyffe, and Noah Snavely.
\newblock Stereo magnification: Learning view synthesis using multiplane images.
\newblock 2018.

\end{thebibliography}
}

\ifarxiv \clearpage \appendix \section{Implementation Details}
\label{sec:appendix_section}
We implement LongSplat using PyTorch. Our rendering and 3D Gaussian updates are accelerated using CUDA and cuDNN. Camera pose optimization is performed using differentiable rendering, while the PnP initialization leverages OpenCV’s solver with RANSAC. All experiments run on NVIDIA 4090 GPUs.

\subsection{LongSplat Algorithm: Pseudo-Code}

The LongSplat pipeline incrementally reconstructs a scene from a casually captured long video, without known poses, by tightly coupling pose estimation and 3D Gaussian Splatting. The workflow can be summarized in the following pseudo-code:

\begin{algorithm}[htbp]\small
\SetAlgoLined
\DontPrintSemicolon

\KwIn{RGB frames $\{I_t\}_{t=1}^{T}$}
\KwOut{3DGS $\mathcal{G}$, camera poses $\{P_t\}_{t=1}^{T}$}

\tcc{\textbf{Initialization}}

$(D_t,C_t, P_t) \leftarrow \texttt{MASt3R Global Alignment}(I_{1...N_{\mathrm{init}}})$\;

\texttt{OctreeAnchorFormation}$(\mathcal{G},D_t,P_t)$\;

\tcc{\textbf{Incremental Joint Optimization}}

\For{$t \leftarrow N_{\mathrm{init}}$ \KwTo $T$}{
    \texttt{GlobalOptimize}$(\mathcal{G},\{P_{1..{t-1}}\},K_g)$\;

    $(D_t,C_t) \leftarrow \texttt{MASt3R}(I_t)$\;

    $P_t \leftarrow \texttt{PnP\_RANSAC}(C_t,\mathcal{G})$\;

    \If{$P_t = \textit{FAIL}$}{
        fallback to $t$
    }

    \texttt{PoseRefine}$(\mathcal{G},P_t,I_t)$\;
    
    \texttt{AnchorUnprojection}$(\mathcal{G},D_t,P_t)$\;
    
    $\mathcal{W}\leftarrow\texttt{VisibilityWindow}(t)$\;
    
    \texttt{LocalOptimize}$(\mathcal{G},\{P_k\}_{k\in\mathcal{W}},K_\ell)$\;

}

\tcc{\textbf{Final Global Refinement}}

\texttt{GlobalRefinement}$(\mathcal{G},\{P_{1..T}\},K_r)$\;

\Return{$(\mathcal{G},\{P_t\}_{t=1}^{T})$}
\caption{\textsc{LongSplat}: Incremental 3DGS}
\end{algorithm}




\section{Additional Experiments}


\subsection{CO3Dv2 Benchmark Evaluation.}
We report the results on CO3Dv2~\cite{reizenstein2021common} in \cref{fig:co3d_visual} and Table~\ref{tab:co3d_average_quantitative}. LongSplat surpasses CF-3DGS and HT-3DGS in all image and pose metrics, confirming the method’s robustness on this more challenging benchmark.

\begin{table}[t]
\centering
\scriptsize
\setlength{\aboverulesep}{0.2pt}
\setlength{\belowrulesep}{0.2pt}
\renewcommand{\arraystretch}{1.0}
\caption{\textbf{Qualtitative comparison on the CO3Dv2 dataset~\cite{reizenstein2021common}}}
\label{tab:co3d_average_quantitative}
\resizebox{1.0\columnwidth}{!} {
\begin{tabular}{l|l|ccc|ccc}
\toprule
Dataset & Method & PSNR $\uparrow$ & SSIM $\uparrow$ & LPIPS $\downarrow$ & ATE$\downarrow$ & RPE$_t$$\downarrow$ & RPE$_r$$\downarrow$ \\
\midrule
\multirow{3}{*}{CO3Dv2}
  & CF-3DGS & 26.61 & 0.79 & 0.29 & 0.014 & 0.218 & 0.374  \\
  & HT-3DGS & 28.34 & 0.84 & 0.30 & 0.017 & 0.058 & 0.314 \\
  & Ours & \textbf{32.59} & \textbf{0.91} & \textbf{0.17} & \textbf{0.005} & \textbf{0.023} & \textbf{0.096}\\
\bottomrule
\end{tabular}
}
\end{table}
\begin{figure*}[t]
\centering
\includegraphics[width=\textwidth]{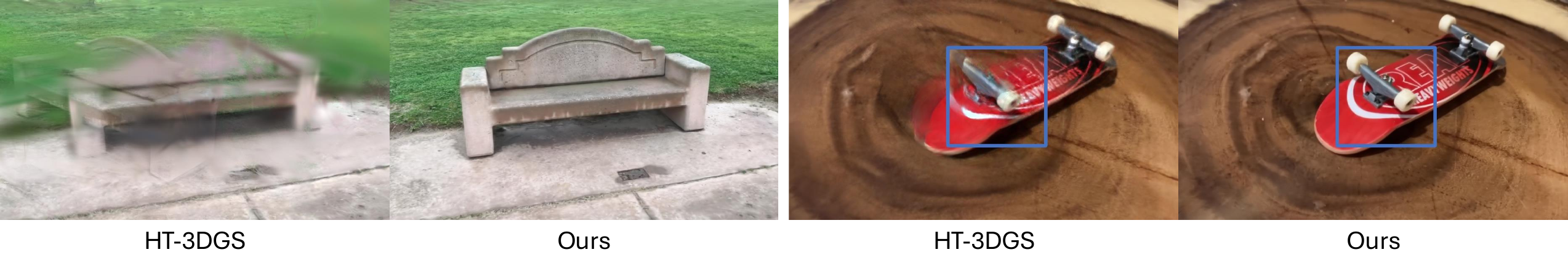}
\caption{\textbf{Qualitative comparison on the CO3Dv2 dataset~\cite{reizenstein2021common}}}
\label{fig:co3d_visual}
\end{figure*}

\subsection{Comparison between COLMAP and LongSplat on the Hike Dataset}
We compare LongSplat with a standard COLMAP-based reconstruction pipeline on our Hike dataset. This dataset poses extreme challenges for incremental SfM due to vegetation occlusion, textureless surfaces, and long trajectories. The quantitative results in Table~\ref{tab:hike_quantitative_colmap} show that LongSplat consistently outperforms COLMAP in both rendering quality and pose estimation accuracy. This highlights the advantage of our octree-anchored Gaussian formulation combined with learned 3D priors.

\subsection{Pose Accuracy on Hike Dataset.}
COLMAP poses are \textbf{noisy} on several Hike videos, so we use the 6 stable sequences (forest2, indoor, university1-4) as references to compute pose accuracy in Table~\ref{tab:hike_pose}. LongSplat achieves the lowest errors, beating all baselines.

\begin{table}[t]
\caption{\textbf{Pose Accuracy on Hike Dataset.}}
\label{tab:hike_pose}
\begin{tabular}{l|ccc}
\toprule
Hike dataset & ATE$\downarrow$ & RPE$_t$$\downarrow$ & RPE$_r$$\downarrow$  \\
\midrule
MASt3R + Scaffold-GS & 0.006 & 0.009 & 0.292  \\
MASt3R + Scaffold-GS* & 0.006 & 0.009 & 0.221  \\
LocalRF & 0.004 & 0.011 & 0.211\\
Ours & \textbf{0.002} & \textbf{0.003} & \textbf{0.128} \\
\bottomrule
\end{tabular}
\end{table}

\subsection{Comparison between HT-3DGS and LongSplat}
We report the comparison with HT-3DGS in Table~\ref{tab:comp_w_ht3dgs} and \cref{fig:ht3dgs_visual}. HT-3DGS runs only on T\&T (33.53 dB), but falls to 13.75 dB on Free and runs OOM on Hike. LongSplat remains stable across all datasets. This confirms our SOTA claim for long, casually captured videos.
\begin{table}[t]
    \centering
    \small
    \renewcommand{\arraystretch}{0.8}
    \caption{\textbf{Qualitative comparison with HT-3DGS.}}
    \label{tab:comp_w_ht3dgs}
    \resizebox{\columnwidth}{!}{
    \begin{tabular}{l|l|ccc|ccccc}
    \toprule
    Dataset & Method & PSNR $\uparrow$ & SSIM $\uparrow$ & LPIPS $\downarrow$ & ATE$\downarrow$ & RPE$_t$$\downarrow$ & RPE$_r$$\downarrow$ & Success Rate \\
    \midrule
    \multirow{2}{*}{Tanks \& Temples}
      & HT-3DGS & \textbf{33.53} & \textbf{0.96} & \textbf{0.07} & \textbf{0.00} & 0.04 & \textbf{0.07} & 8/8 \\
      & Ours & 32.83 & 0.94 & 0.08 & \textbf{0.00} & \textbf{0.03} & \textbf{0.07} & 8/8 \\
    \midrule
    \multirow{2}{*}{Free}
      & HT-3DGS & 13.75 & 0.39 & 0.65 & 0.02 & 0.34 & 4.41 & 6/7 \\
      & Ours & \textbf{27.88} & \textbf{0.85} & \textbf{0.17} & \textbf{0.00} & \textbf{0.03} &\textbf{0.10} & 7/7 \\
    \midrule
    \multirow{2}{*}{Hike}
      & HT-3DGS & OOM & OOM & OOM & OOM & OOM & OOM & 0/12 \\
      & Ours & \textbf{25.39} & \textbf{0.81} & \textbf{0.19} & \textbf{0.00} & \textbf{0.01} & \textbf{0.21} & 12/12 \\
    \bottomrule
    \end{tabular}}
\end{table}

\begin{figure*}[t]
\centering
\includegraphics[width=\textwidth]{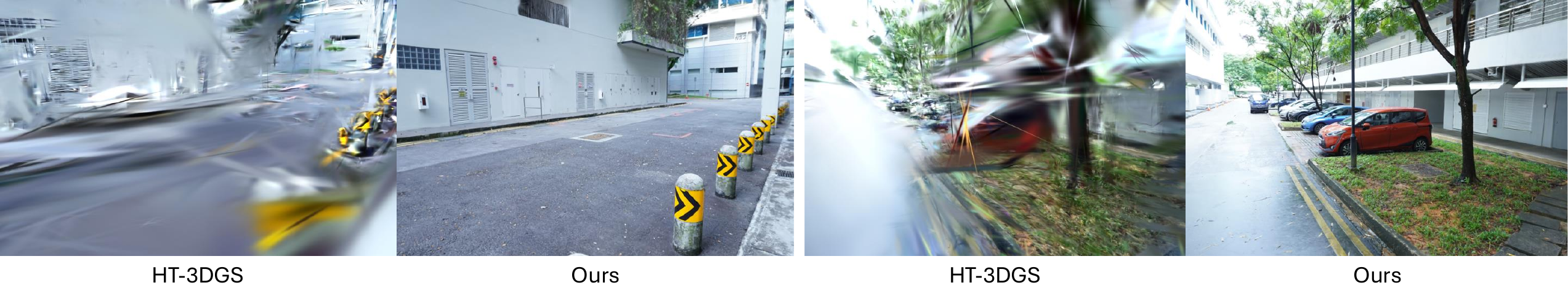}
\caption{\textbf{Qualitative comparison with HT-3DGS}}
\label{fig:ht3dgs_visual}
\end{figure*}

\subsection{Ablation on Using MASt3R Relative Poses}
\begin{figure*}[t]
\centering
\resizebox{1.0\textwidth}{!} 
{
\includegraphics[width=\textwidth]{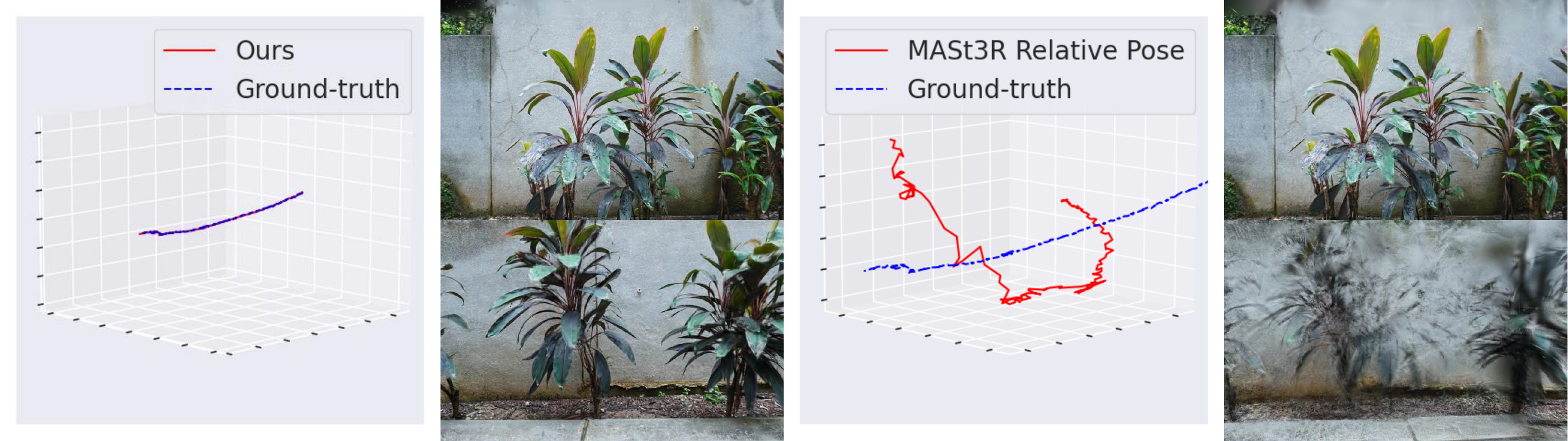}
}
\vspace{-6mm}
\caption{\textbf{Visual comparisons on ablation MASt3R relative pose.}}
\label{fig:ablation_mast3r}
\end{figure*}

To demonstrate the importance of our proposed pose estimation pipeline, we conduct an ablation replacing LongSplat’s correspondence-guided PnP with directly using MASt3R’s relative pose estimates. As shown in \cref{fig:ablation_mast3r}, this leads to degraded novel view synthesis quality and larger pose errors, especially in long sequences. This confirms that raw MASt3R poses alone are insufficient for high-quality incremental reconstruction.

\subsection{Ablation on training loss}
We report the ablation study on training loss in Table~\ref{tab:abalation_training_loss}. Removing individual losses degrades performance. Our full method achieves the best rendering quality and pose accuracy.

\begin{table}[t]
    \centering
    \small
    \renewcommand{\arraystretch}{0.8}
    \caption{\textbf{Ablation on training loss.}}
    \label{tab:abalation_training_loss}
    \resizebox{\columnwidth}{!}{
    \begin{tabular}{l|cccccc}
    \toprule
    Method & PSNR$\uparrow$ & SSIM$\uparrow$ & LPIPS$\downarrow$ & RPE$_t$$\downarrow$ & RPE$_r$$\downarrow$ & ATE$\downarrow$ \\
    \midrule
    w/o 2d correspondence loss & 26.54 & 0.80 & 0.24 & 0.049 & 0.253 & 0.007 \\
    w/o depth loss             & 26.74 & 0.82 & 0.22 & 0.076 & 0.246 & 0.011 \\
    Ours                       & \textbf{27.88} & \textbf{0.85} & \textbf{0.17} & \textbf{0.028} & \textbf{0.103} & \textbf{0.004} \\
    \bottomrule
    \end{tabular}%
    }
\end{table}

\begin{table*}[t]
\centering
\caption{\textbf{Quantitative evaluation on the Hike dataset~\cite{meuleman2023progressively}.} Our method consistently outperforms baselines across diverse scenes with complex trajectories and extended sequences, highlighting LongSplat’s robustness and superior scene representation capability. CF-3DGS~\cite{fu2024colmap} encounters OOM in all scenes and is thus omitted.}
\label{tab:hike_quantitative_colmap}
\footnotesize
\setlength{\tabcolsep}{1pt}
\begin{tabular}{l|ccc ccc ccc ccc ccc}
\toprule
\multirow{3}{*}{Scenes} 
  & \multicolumn{3}{c}{COLMAP} 
  & \multicolumn{3}{c}{MASt3R~\cite{leroy2024grounding}} 
  & \multicolumn{3}{c}{MASt3R~\cite{leroy2024grounding}} 
  & \multicolumn{3}{c}{\multirow{2}{*}{LocalRF~\cite{meuleman2023progressively}}} 
  & \multicolumn{3}{c}{\multirow{2}{*}{Ours}} \\
  & \multicolumn{3}{c}{+ Scaffold-GS~\cite{lu2024scaffold}} 
  & \multicolumn{3}{c}{+ Scaffold-GS~\cite{lu2024scaffold}} 
  & \multicolumn{3}{c}{+ Scaffold-GS~\cite{lu2024scaffold}} 
  & & & 
  & & \\[1ex]
\cmidrule(lr){2-4} \cmidrule(lr){5-7} \cmidrule(lr){8-10} \cmidrule(lr){11-13} \cmidrule(lr){14-16}
  & PSNR$\uparrow$ & SSIM$\uparrow$ & LPIPS$\downarrow$ 
  & PSNR$\uparrow$ & SSIM$\uparrow$ & LPIPS$\downarrow$ 
  & PSNR$\uparrow$ & SSIM$\uparrow$ & LPIPS$\downarrow$ 
  & PSNR$\uparrow$ & SSIM$\uparrow$ & LPIPS$\downarrow$ 
  & PSNR$\uparrow$ & SSIM$\uparrow$ & LPIPS$\downarrow$ \\
\midrule
forest1     & 20.12 & 0.55 & 0.44 & 17.68 & 0.30 & 0.64 & 17.54 & 0.34 & 0.55 & 19.12 & 0.45 & 0.41 & 23.86 & 0.79 & 0.21 \\
forest2     & 28.35 & 0.89 & 0.14 & 20.91 & 0.53 & 0.36 & 21.11 & 0.54 & 0.35 & 27.23 & 0.84 & 0.15 & 27.87 & 0.88 & 0.11 \\
forest3     & - & - & - & 9.54  & 0.15 & 0.70 & 9.62  & 0.15 & 0.70 & 17.05 & 0.38 & 0.59 & 19.59 & 0.62 & 0.31 \\
garden1     & 20.77 & 0.67 & 0.28 & 13.09 & 0.23 & 0.75 & 14.84 & 0.27 & 0.72 & 22.11 & 0.66 & 0.28 & 24.12 & 0.80 & 0.19 \\
garden2     & -     & -    & -    & 13.21 & 0.19 & 0.75 & 15.67 & 0.26 & 0.74 & 23.34 & 0.61 & 0.33 & 24.35 & 0.74 & 0.25 \\
garden3     & 23.46 & 0.73 & 0.23 & 11.82 & 0.13 & 0.64 & 11.89 & 0.13 & 0.64 & 23.33 & 0.67 & 0.27 & 24.01 & 0.75 & 0.23 \\
indoor      & 28.85 & 0.90 & 0.19 & 23.64 & 0.81 & 0.33 & 24.64 & 0.83 & 0.31 & 30.17 & 0.91 & 0.17 & 30.62 & 0.92 & 0.17 \\
playground  & -     & -    & -    & 19.31 & 0.49 & 0.40 & 19.73 & 0.52 & 0.38 & 22.29 & 0.63 & 0.28 & 24.30 & 0.78 & 0.18 \\
university1 & 25.36 & 0.78 & 0.27 & 19.38 & 0.47 & 0.53 & 19.62 & 0.48 & 0.52 & 25.22 & 0.71 & 0.32 & 25.50 & 0.79 & 0.24 \\
university2 & 27.25 & 0.87 & 0.13 & 20.27 & 0.58 & 0.36 & 20.72 & 0.60 & 0.35 & 24.56 & 0.75 & 0.23 & 26.82 & 0.85 & 0.15 \\
university3 & 26.98 & 0.89 & 0.13 & 18.59 & 0.51 & 0.39 & 19.31 & 0.57 & 0.35 & 23.23 & 0.73 & 0.23 & 25.57 & 0.86 & 0.13 \\
university4 & 25.03 & 0.82 & 0.17 & 20.23 & 0.61 & 0.39 & 20.13 & 0.61 & 0.39 & 25.08 & 0.79 & 0.22 & 27.00 & 0.88 & 0.12 \\
\midrule
Avg         & 25.13 & 0.79 & 0.22 & 17.30 & 0.42 & 0.52 & 17.90 & 0.44 & 0.50 & 23.56 & 0.68 & 0.29 & \textbf{25.39} & \textbf{0.81} & \textbf{0.19} \\
\bottomrule
\end{tabular}
\end{table*}

\section{Complete Quantitative Evaluation}

\subsection{Tanks and Temples}
\begin{table*}[t]
\caption{\textbf{Quantitative evaluation of novel view synthesis quality on the Tanks and Temples dataset~\cite{knapitsch2017tanks}.} Our proposed LongSplat consistently surpasses existing methods across multiple challenging scenes.}
\label{tab:tnt_quantitative}
\centering
\footnotesize
\setlength{\tabcolsep}{1pt}
\begin{tabular}{l|ccc ccc ccc ccc}
\toprule
\multirow{3}{*}{Scenes} & \multicolumn{3}{c}{COLMAP+3DGS~\cite{kerbl20233d}} & \multicolumn{3}{c}{NoPe-NeRF~\cite{bian2023nope}} & \multicolumn{3}{c}{CF-3DGS~\cite{fu2024colmap}} & \multicolumn{3}{c}{Ours} \\
\cmidrule(lr){2-4} \cmidrule(lr){5-7} \cmidrule(lr){8-10} \cmidrule(lr){11-13}
& PSNR$\uparrow$ & SSIM$\uparrow$ & LPIPS$\downarrow$ & PSNR$\uparrow$ & SSIM$\uparrow$ & LPIPS$\downarrow$ & PSNR$\uparrow$ & SSIM$\uparrow$ & LPIPS$\downarrow$ & PSNR$\uparrow$ & SSIM$\uparrow$ & LPIPS$\downarrow$ \\
\midrule
Church    & 29.93 & 0.93 & 0.09 & 25.17 & 0.73 & 0.39 & 30.23 & 0.93 & 0.11 & 30.96 & 0.93 & 0.10 \\
Barn      & 31.08 & 0.95 & 0.07 & 26.35 & 0.69 & 0.44 & 31.23 & 0.90 & 0.10 & 32.57 & 0.92 & 0.09 \\
Museum    & 34.47 & 0.96 & 0.05 & 26.77 & 0.76 & 0.35 & 29.91 & 0.91 & 0.11 & 33.78 & 0.95 & 0.06 \\
Family    & 27.93 & 0.92 & 0.11 & 26.01 & 0.74 & 0.41 & 31.27 & 0.94 & 0.07 & 33.67 & 0.96 & 0.06 \\
Horse     & 20.91 & 0.77 & 0.23 & 27.64 & 0.84 & 0.26 & 33.94 & 0.96 & 0.05 & 33.42 & 0.96 & 0.06 \\
Ballroom  & 34.48 & 0.96 & 0.04 & 25.33 & 0.72 & 0.38 & 32.47 & 0.96 & 0.07 & 32.80 & 0.95 & 0.06 \\
Francis   & 32.64 & 0.92 & 0.15 & 29.48 & 0.80 & 0.38 & 32.72 & 0.91 & 0.14 & 33.80 & 0.92 & 0.15 \\
Ignatius  & 30.20 & 0.93 & 0.08 & 23.96 & 0.61 & 0.47 & 28.43 & 0.90 & 0.09 & 31.61 & 0.94 & 0.07 \\
\midrule
Avg.      & 30.21 & 0.92 & 0.10 & 26.34 & 0.74 & 0.39 & 31.28 & 0.93 & 0.09 & \textbf{32.83} & \textbf{0.94} & \textbf{0.08} \\
\bottomrule
\end{tabular}
\end{table*}

\begin{table*}[t]
\caption{\textbf{Quantitative evaluation of camera pose estimation accuracy on the Tanks and Temples dataset~\cite{knapitsch2017tanks}.} Our method achieves consistently low errors across diverse scenes, outperforming CF-3DGS and NoPe-NeRF, especially in terms of global trajectory accuracy (ATE) and local translation consistency ($\text{RPE}_t$).}
\label{tab:tnt_pose_quantitative}
\centering
\footnotesize
\setlength{\tabcolsep}{2pt}
\begin{tabular}{l|ccc ccc ccc}
\toprule
\multirow{3}{*}{Scenes} & \multicolumn{3}{c}{CF-3DGS} & \multicolumn{3}{c}{NoPe-NeRF} & \multicolumn{3}{c}{Ours} \\
\cmidrule(lr){2-4} \cmidrule(lr){5-7} \cmidrule(lr){8-10} 
& ATE$\downarrow$ & $\text{RPE}_r$$\downarrow$ & $\text{RPE}_t$$\downarrow$ & ATE$\downarrow$ & $\text{RPE}_r$$\downarrow$ & $\text{RPE}_t$$\downarrow$ & ATE$\downarrow$ & $\text{RPE}_r$$\downarrow$ & $\text{RPE}_t$$\downarrow$ \\
\midrule
Church    & 0.002 & 0.018 & 0.008 & 0.008 & 0.008 & 0.034 & 0.001 & 0.048 & 0.011 \\
Barn      & 0.003 & 0.034 & 0.034 & 0.004 & 0.032 & 0.046 & 0.004 & 0.061 & 0.025 \\
Museum    & 0.005 & 0.215 & 0.052 & 0.020 & 0.202 & 0.207 & 0.001 & 0.046 & 0.025 \\
Family    & 0.002 & 0.024 & 0.022 & 0.001 & 0.015 & 0.047 & 0.002 & 0.043 & 0.021 \\
Horse     & 0.003 & 0.057 & 0.112 & 0.003 & 0.017 & 0.179 & 0.001 & 0.046 & 0.086 \\
Ballroom  & 0.003 & 0.024 & 0.037 & 0.002 & 0.018 & 0.041 & 0.002 & 0.053 & 0.021 \\
Francis   & 0.006 & 0.154 & 0.029 & 0.005 & 0.009 & 0.057 & 0.009 & 0.213 & 0.036 \\
Ignatius  & 0.005 & 0.032 & 0.033 & 0.002 & 0.005 & 0.026 & 0.002 & 0.034 & 0.032 \\
\midrule
Avg.      & 0.004 & 0.069 & 0.041 & 0.006 & \textbf{0.038} & 0.080 & \textbf{0.003} & 0.068 & \textbf{0.032} \\
\bottomrule
\end{tabular}
\end{table*}

We provide full quantitative results on the Tanks and Temples benchmark in \cref{tab:tnt_quantitative,tab:tnt_pose_quantitative}. LongSplat consistently outperforms baselines in both rendering quality and pose estimation accuracy, demonstrating its effectiveness even in indoor and urban scenes with varied scales and complexities.

\subsection{Free dataset}
\begin{table*}[t]
\caption{\textbf{Quantitative evaluation of camera pose estimation accuracy on the Free dataset~\cite{wang2023f2}.} ``-'' indicates methods that encountered out-of-memory issues. Our method consistently achieves superior performance across most scenes, significantly reducing pose errors compared to state-of-the-art approaches. ``*'': Initialized with MASt3R poses, then jointly optimized.}
\label{tab:free_pose_quantitative_full}
\centering
\footnotesize
\resizebox{\textwidth}{!}{%
\setlength{\tabcolsep}{2pt}
\begin{tabular}{l|ccc ccc|ccc ccc ccc ccc}
\toprule
\multirow{3}{*}{Scenes} & \multicolumn{3}{c}{MASt3R~\cite{leroy2024grounding}} & \multicolumn{3}{c|}{MASt3R~\cite{leroy2024grounding}} & \multicolumn{3}{c}{\multirow{2}{*}{CF-3DGS~\cite{fu2024colmap}}} & \multicolumn{3}{c}{\multirow{2}{*}{NoPe-NeRF~\cite{bian2023nope}}} & \multicolumn{3}{c}{\multirow{2}{*}{LocalRF~\cite{meuleman2023progressively}}} & \multicolumn{3}{c}{\multirow{2}{*}{Ours}} \\
& \multicolumn{3}{c}{+ Scaffold-GS~\cite{lu2024scaffold}} & \multicolumn{3}{c|}{+ Scaffold-GS~\cite{lu2024scaffold}*} & & & & & & & & & & & \\
\cmidrule(lr){2-4} \cmidrule(lr){5-7} \cmidrule(lr){8-10} \cmidrule(lr){11-13} \cmidrule(lr){14-16} \cmidrule(lr){17-19} 
& ATE $\downarrow$ & $\text{RPE}_r$ $\downarrow$ & $\text{RPE}_t$ $\downarrow$ & ATE $\downarrow$ & $\text{RPE}_r$ $\downarrow$ & $\text{RPE}_t$ $\downarrow$ & ATE $\downarrow$ & $\text{RPE}_r$ $\downarrow$ & $\text{RPE}_t$ $\downarrow$ & ATE $\downarrow$ & $\text{RPE}_r$ $\downarrow$ & $\text{RPE}_t$ $\downarrow$ & ATE $\downarrow$ & $\text{RPE}_r$ $\downarrow$ & $\text{RPE}_t$ $\downarrow$ & ATE $\downarrow$ & $\text{RPE}_r$ $\downarrow$ & $\text{RPE}_t$ $\downarrow$  \\
\midrule
Grass    & 0.038 & 0.554 & 0.559 & 0.002 & 0.152 & 0.016 & -   & -   & -   & 0.431 & 9.333 & 3.044 & 0.056 & 6.026 & 0.612 & 0.000 & 0.058 & 0.002 \\
Hydrant  & 0.013 & 0.168 & 0.145 & 0.013 & 0.165 & 0.144 & -   & -   & -   & 0.480 & 4.068 & 5.844 & 0.060 & 8.487 & 1.068 & 0.013 & 0.111 & 0.069 \\
Lab      & 0.009 & 0.294 & 0.175 & 0.009 & 0.265 & 0.178 & -   & -   & -   & 0.533 & 2.623 & 5.774 & 0.041 & 4.405 & 1.072 & 0.004 & 0.217 & 0.067 \\
Pillar   & 0.003 & 0.225 & 0.024 & 0.003 & 0.199 & 0.016 & 0.023 & 4.744 & 0.328 & 0.576 & 4.176 & 2.013 & 0.025 & 3.553 & 0.526 & 0.001 & 0.066 & 0.003 \\
Road     & 0.013 & 0.153 & 0.088 & 0.013 & 0.159 & 0.088 & -   & -   & -   & 0.584 & 4.087 & 6.045 & 0.023 & 9.798 & 0.699 & 0.005 & 0.080 & 0.036 \\
Sky      & 0.010 & 0.203 & 0.091 & 0.010 & 0.197 & 0.090 & - & - & - & 0.807 & 6.661 & 9.775 & 0.031 & 11.075 & 0.894 & 0.002 & 0.114 & 0.017 \\
Stair    & 0.006 & 0.260 & 0.050 & 0.006 & 0.247 & 0.050 & 0.021 & 2.139 & 0.140 & 0.624 & 2.809 & 11.120 & 0.008 & 6.257 & 0.563 & 0.000 & 0.078 & 0.001 \\
\midrule
Avg.      & 0.013 & 0.265 & 0.162 & 0.008 & 0.198 & 0.083 & 0.019 & 4.365 & 0.290 & 0.576 & 4.822 & 6.231 & 0.035 & 7.086 & 0.776 & \textbf{0.004} & \textbf{0.103} & \textbf{0.028} \\
\bottomrule
\end{tabular}
}
\end{table*}
We provide full quantitative results on the Free dataset benchmark in \cref{tab:free_pose_quantitative_full}. LongSplat consistently outperforms baselines in both rendering quality and pose estimation accuracy, demonstrating its effectiveness even in indoor and urban scenes with varied scales and complexities.

\subsection{Hike dataset}
Hike dataset benchmark in \cref{tab:hike_quantitative_colmap}. LongSplat consistently outperforms baselines in both rendering quality and pose estimation accuracy, demonstrating its effectiveness even in challenging indoor and urban scenes with varied scales and complexities. Notably, in scenarios where COLMAP fails to reconstruct due to long trajectories or low-texture regions, LongSplat maintains high-quality results, preserving structural details and ensuring stable pose estimation.

\section{Additional Visual Comparisons}

\subsection{Visual Comparison on Ablation Study}
\begin{figure*}[t]
\centering
\resizebox{1.0\textwidth}{!} 
{
\includegraphics[width=\textwidth]{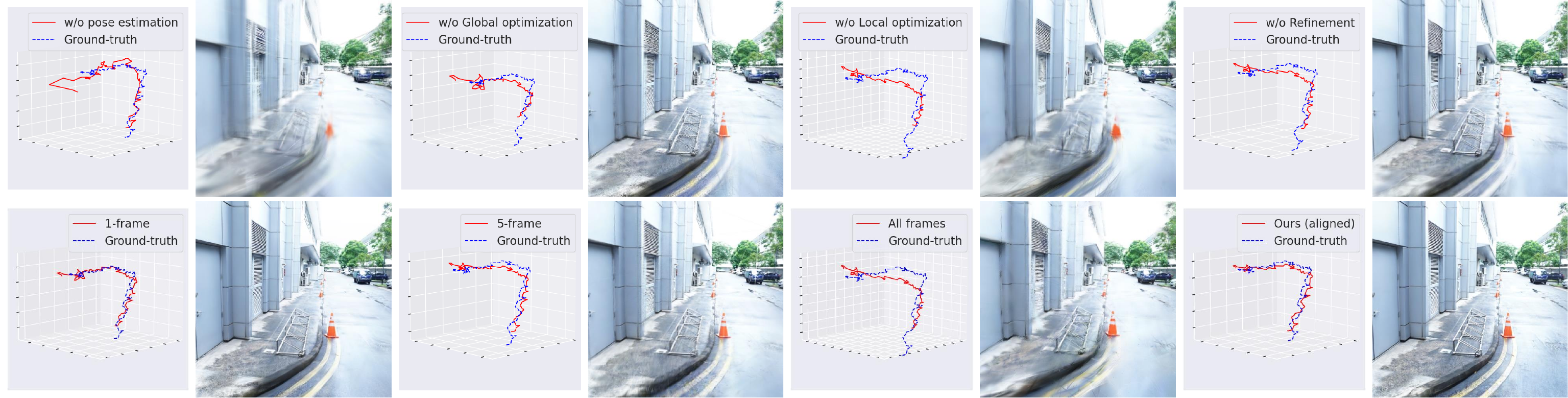}
}
\caption{\textbf{Visual comparisons on ablation studies.} 
The top row shows the camera trajectory estimation and novel view synthesis results when different training components are removed, demonstrating the importance of each proposed module. Removing global optimization, local optimization, or final refinement significantly degrades pose accuracy and reconstruction quality. The bottom row evaluates different settings for the visibility-adapted local window size. Too small a window leads to unstable geometry and pose drift, while too large a window dilutes local visibility priors, slowing convergence. LongSplat achieves the best balance using the proposed adaptive window.}
\label{fig:ablation_training}
\end{figure*}

\cref{fig:ablation_training} shows the visual impact of removing key training components. Both trajectory estimation and novel view synthesis degrade severely when global optimization, local optimization, or final refinement is removed, emphasizing their importance.


\subsection{Additional Trajectory Results}
\begin{figure*}[t]
\centering
{
\includegraphics[width=\textwidth]{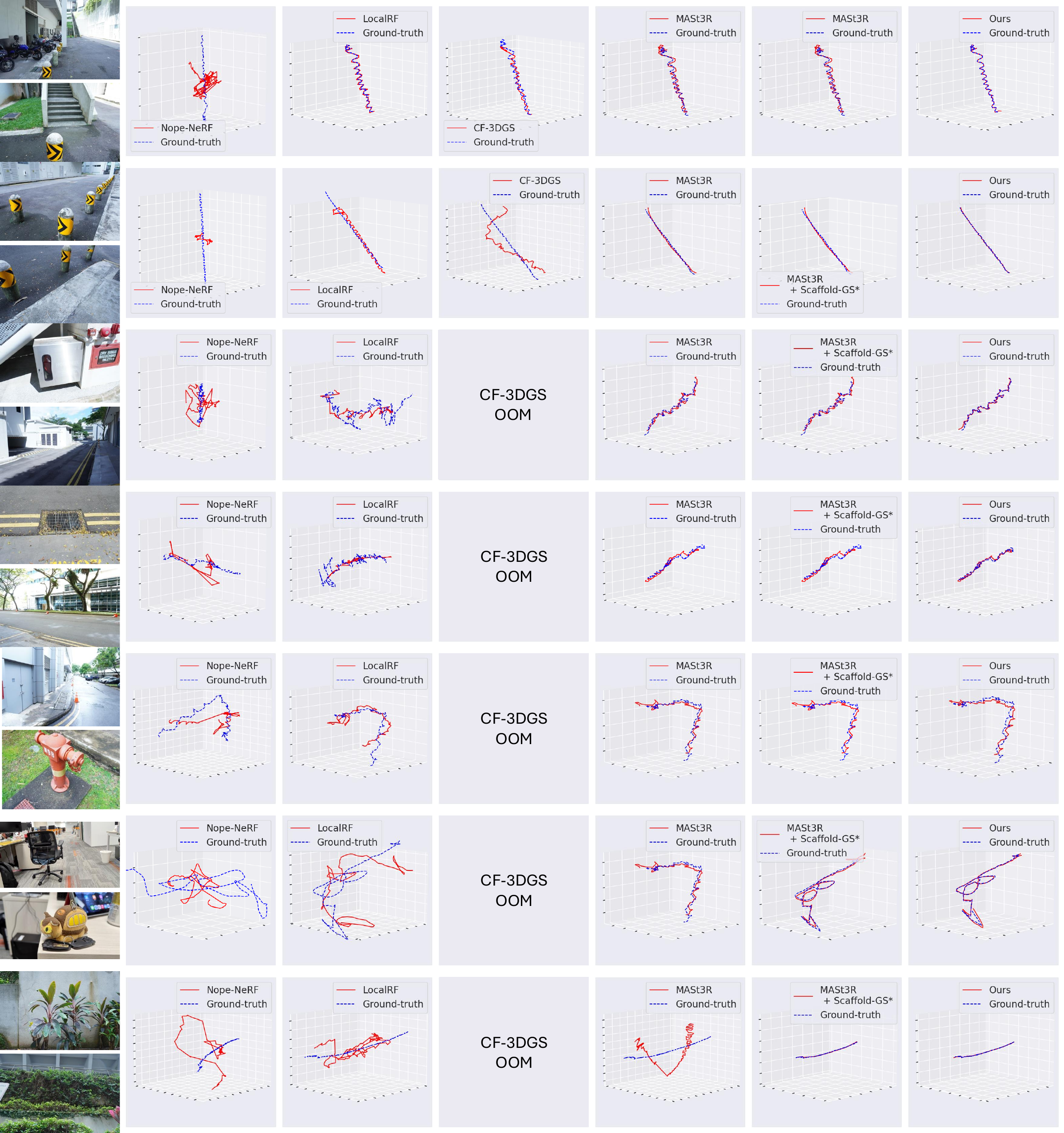}
}

\caption{
\textbf{Visualization of camera trajectories on Free dataset~\cite{wang2023f2}.} CF-3DGS~\cite{fu2024colmap} encounters OOM and fails for long sequences, whereas our method reliably estimates accurate, stable trajectories, demonstrating superior robustness.
}
\label{fig:free_traj_more_visual}
\end{figure*}

We include additional visualizations of camera trajectories estimated by LongSplat. As shown in \cref{fig:free_traj_more_visual}, our method reconstructs stable, drift-free trajectories even in long and complex sequences.

\subsection{Additional Tanks and Temples Results}
\begin{figure*}[t]
\centering
\resizebox{1.0\textwidth}{!} 
{
\includegraphics[width=\textwidth]{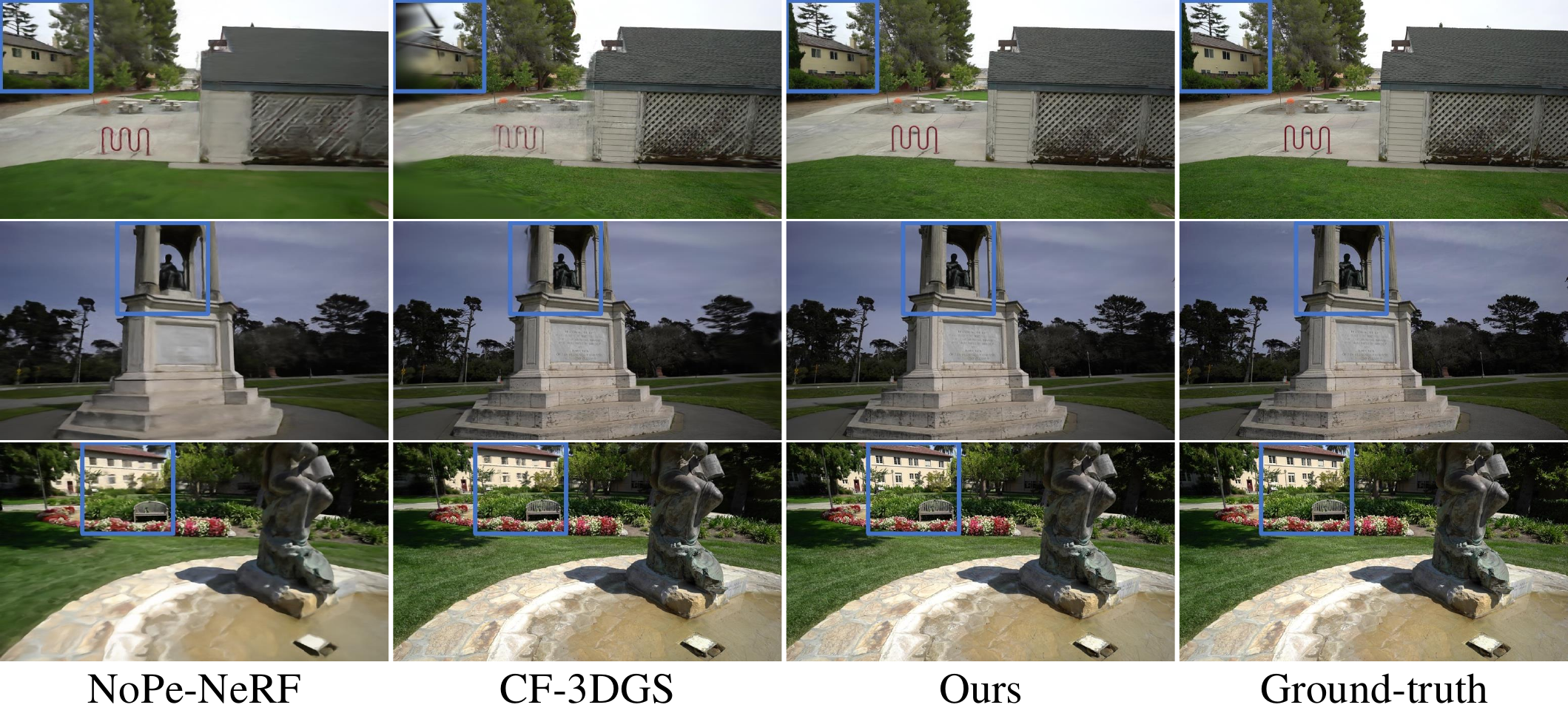}
}
\caption{\textbf{More Qualitative comparison on the Tanks and Temples dataset~\cite{knapitsch2017tanks}.} NoPe-NeRF~\cite{bian2023nope} produces visibly blurred results with inaccurate geometries, while CF-3DGS~\cite{fu2024colmap}, despite better sharpness, fails to reconstruct fine details accurately. In contrast, our LongSplat method achieves superior rendering quality, closely matching the ground truth with sharper textures, more accurate geometry, and consistent lighting.}
\label{fig:tnt_visual_more}
\end{figure*}
We provide additional qualitative comparisons on the Tanks and Temples benchmark. LongSplat produces sharper and more visually consistent results across diverse scenes, demonstrating strong generalization across both indoor and outdoor environments.

\subsection{Additional Free Dataset Results}
\begin{figure*}[t]
\centering
\resizebox{1.0\textwidth}{!} 
{
\includegraphics[width=\textwidth]{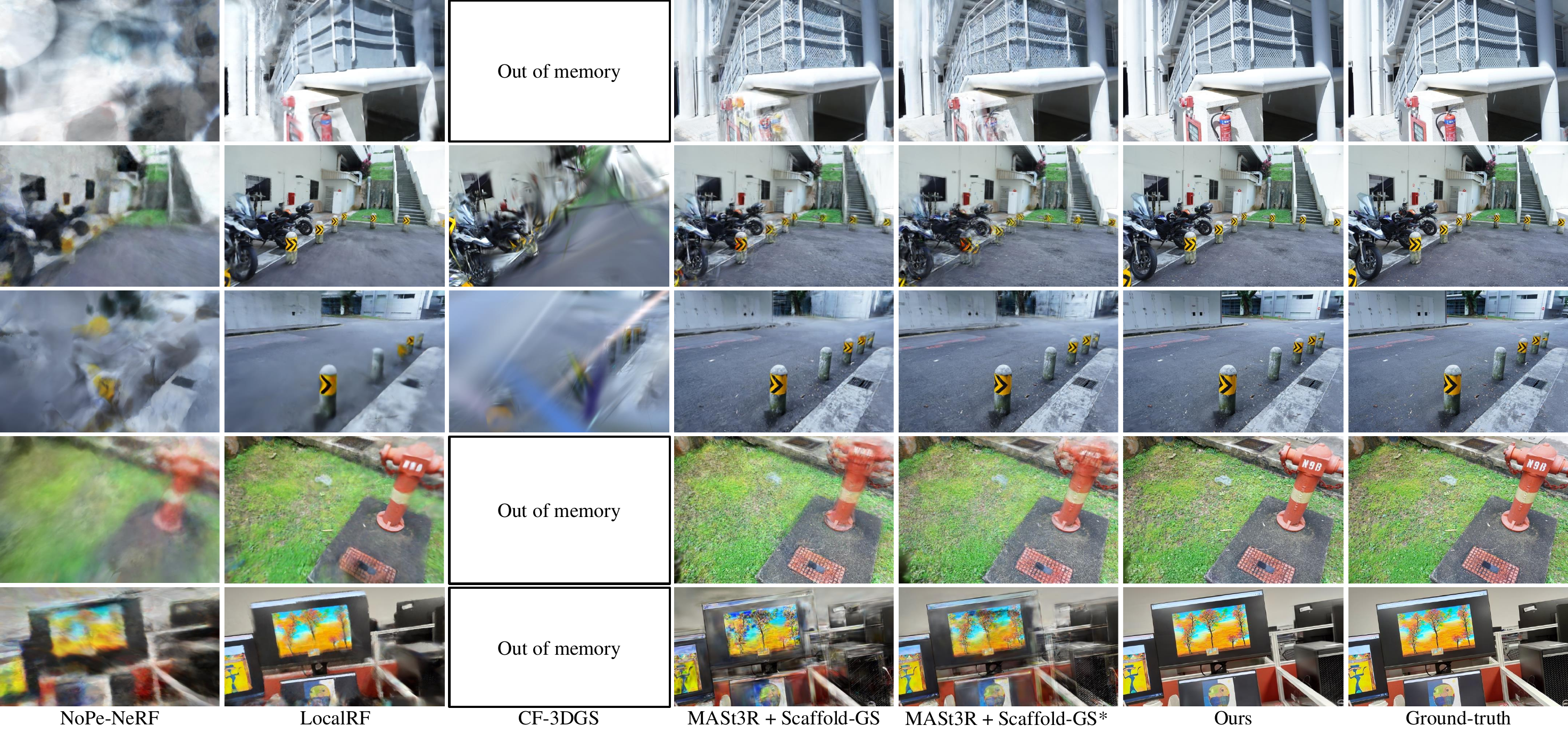}
}
\caption{\textbf{More Qualitative comparison on the Free dataset~\cite{wang2023f2}.} We compare our method with state-of-the-art approaches including NoPe-NeRF~\cite{bian2023nope}, LocalRF~\cite{meuleman2023progressively}, CF-3DGS~\cite{fu2024colmap}, and MASt3R~\cite{leroy2024grounding} combined with Scaffold-GS~\cite{lu2024scaffold}. CF-3DGS fails due to memory constraints (OOM), and other baseline methods exhibit artifacts or blurry reconstructions. In contrast, our method produces results closest to the ground truth, demonstrating clearer details, accurate geometry, and visually consistent rendering, particularly under challenging scene structures and complex camera trajectories. ``*'': Initialized with MASt3R poses, then jointly optimized.}
\label{fig:free_visual_more}
\end{figure*}
Additional qualitative comparisons on the Free dataset are shown in \cref{fig:free_visual_more}. Our method preserves more fine details, produces fewer artifacts, and achieves sharper novel view synthesis than all baselines.

\subsection{Additional Hike Dataset Results}
\begin{figure*}[t]
\centering
\resizebox{1.0\textwidth}{!} 
{
\includegraphics[width=\textwidth]{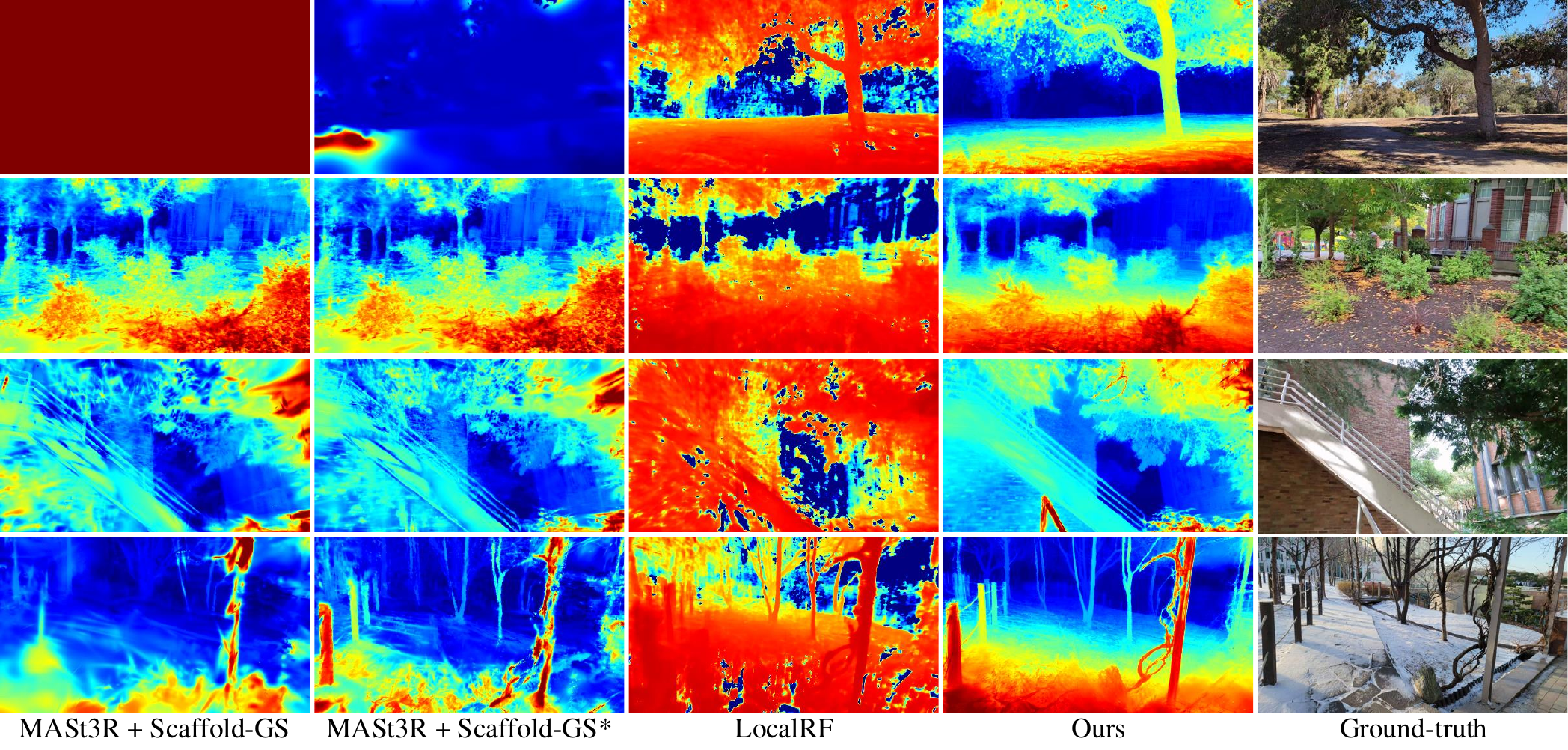}
}
\caption{\textbf{Qualitative results on the Hike dataset~\cite{meuleman2023progressively}.} Compared to existing methods such as LocalRF~\cite{meuleman2023progressively} and MASt3R~\cite{leroy2024grounding}+Scaffold-GS~\cite{lu2024scaffold}, our approach significantly improves visual clarity and reconstruction fidelity, accurately capturing complex details and textures in challenging scenes captured during long, casual outdoor trajectories. Notably, our method better preserves structural details and reduces artifacts, demonstrating enhanced robustness and visual quality. ``*'': Initialized with MASt3R poses, then jointly optimized.}
\label{fig:hike_visual_depth}
\end{figure*}
\begin{figure*}[t]
\centering
\resizebox{1.0\textwidth}{!} 
{
\includegraphics[width=\textwidth]{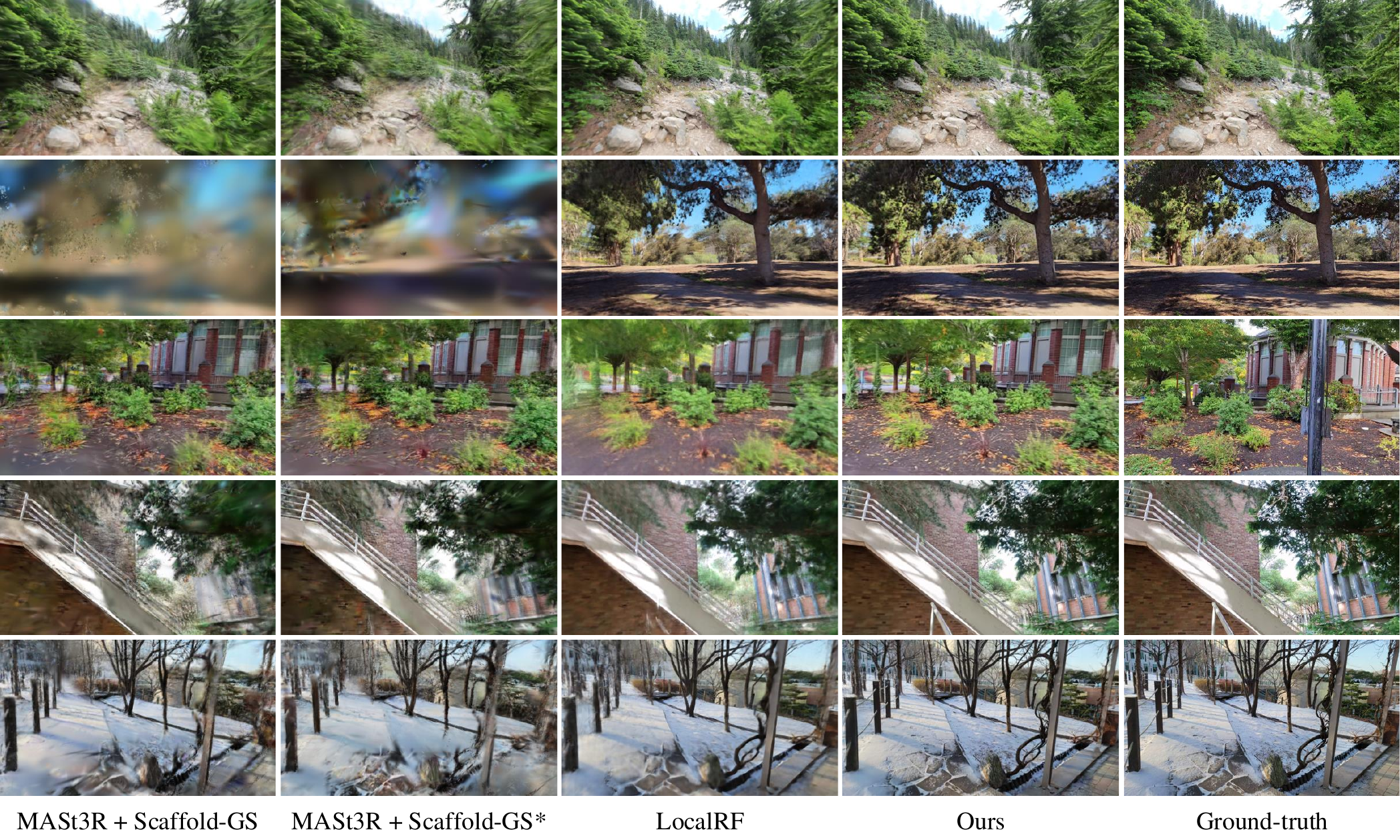}
}
\caption{\textbf{More Qualitative results on the Hike dataset~\cite{meuleman2023progressively}.} Compared to existing methods such as LocalRF~\cite{meuleman2023progressively} and MASt3R~\cite{leroy2024grounding}+Scaffold-GS~\cite{lu2024scaffold}, our approach significantly improves visual clarity and reconstruction fidelity, accurately capturing complex details and textures in challenging scenes captured during long, casual outdoor trajectories. Notably, our method better preserves structural details and reduces artifacts, demonstrating enhanced robustness and visual quality. ``*'': Initialized with MASt3R poses, then jointly optimized.}
\label{fig:hike_visual_more}
\end{figure*}
Finally, we present more qualitative results on the Hike dataset in \cref{fig:hike_visual_depth}, \cref{fig:hike_visual_more}. LongSplat reconstructs complex natural scenes with higher visual quality, capturing vegetation, terrain, and large-scale geometry with remarkable accuracy.

 \fi

\end{document}